\pdfoutput=1

\documentclass[11pt]{article}

\usepackage{emnlp2021}

\usepackage{times}
\usepackage{latexsym}
\usepackage[T1]{fontenc}
\usepackage[utf8]{inputenc}
\usepackage{microtype}

\usepackage{graphicx}
\usepackage{booktabs}
\usepackage{amsthm,amsmath,amssymb,amssymb,bm,mathrsfs}
\usepackage{textcomp}
\usepackage{xcolor}
\usepackage{changes}
\usepackage{algorithm}
\usepackage[noend]{algpseudocode}

\usepackage{listings}
\title{Generalization in Text-based Games via \\ Hierarchical Reinforcement Learning}

\author{Yunqiu Xu$^1$, Meng Fang$^2$, Ling Chen$^1$, Yali Du$^3$, Chengqi Zhang$^1$ \\
  $^1$Australian Artificial Intelligence Institute, University of Technology Sydney, Sydney, Australia \\
  \texttt{Yunqiu.Xu@student.uts.edu.au, \{Ling.Chen,Chengqi.Zhang\}@uts.edu.au}\\
  $^2$Eindhoven University of Technology, Eindhoven, the Netherlands \\
  \texttt{m.fang@tue.nl} \\
  $^3$King’s College London, London, United Kingdom \\
  \texttt{yali.du@kcl.ac.uk} \\}

\begin{document}
\maketitle
\begin{abstract}
Deep reinforcement learning provides a promising approach for text-based games in studying natural language communication between humans and artificial agents. 
However, the generalization still remains a big challenge as the agents depend critically on the complexity and variety of training tasks. 
In this paper, we address this problem by introducing a hierarchical framework built upon the knowledge graph-based RL agent. 
In the high level, a meta-policy is executed to decompose the whole game into a set of subtasks specified by textual goals, and select one of them based on the KG. 
Then a sub-policy in the low level is executed to conduct goal-conditioned reinforcement learning.  
We carry out experiments on games with various difficulty levels and show that the proposed method enjoys favorable generalizability. 
\end{abstract}

\section{Introduction}

Text-based games are simulated systems where the agent takes textual observations as the input, and interacts with the environment via text commands~\cite{hausknecht2019jericho}. 
They are suitable test-beds to study natural language understanding, commonsense reasoning and language-informed decision making~\cite{luketina2019survey}.
Reinforcement Learning (RL) based agents~\cite{narasimhan2015language,zahavy2018nips} have been developed to handle challenges such as language-based representation learning and combinatorial action space. 
Among them, KG-based agents~\cite{ammanabrolu2019kga2c} yield promising performance with the aid of Knowledge Graph (KG), which serves as a belief state to provide structural information.

To design intelligent RL-based agents for text-based games, it is necessary to build agents that automatically learn to solve different games.
However, generalization remains as one of the key challenges of RL $-$ the agent tends to overfit the training environment and fails to generalize to new environments~\cite{cobbe2019rlgeneralization}. 
In the domain of text-based games, the TextWorld~\cite{cote2018textworld} makes it feasible to study generalizability by creating non-overlapping game sets with customizable domain gaps (e.g., themes, vocabulary sets, difficulty levels and layouts).
Most previous works study generalizability either upon games with the same difficulty level but different layouts~\cite{ammanabrolu2019kgdqn}, or upon games with a set of multiple levels that have been observed during training~\cite{adolphs2019ledeepchef}. 
Although these agents perform well on relatively simple games, they can hardly achieve satisfactory performance on difficult games ~\cite{adhikari2020gatav2}.
In this work, we aim to develop agents that can be generalized to not only games from the same difficulty level while having unseen different layouts, but also games from unseen difficulty levels where both layouts and complexities are different.

While solving a whole game might be difficult due to long-term temporal dependencies, and the learnt strategy might be difficult to be transferred to other games due to large domain gaps, it would be more flexible to treat the game as a sketch of subtasks~\cite{andreas2017modular,oh2017zero}.
This brings two branches of benefits. 
First, the subtasks would be easier to solve as they have short-term temporal dependencies. 
Second, the strategies learnt for solving subtasks may be recomposed to solve an unseen game.
Motivated by these insights, we aim to solve a game by decomposing it into subtasks characterized by textual goals, then making decisions conditioned on them.
Instead of hand-crafting the task sketches, we leverage the hierarchical reinforcement learning (HRL) framework for adaptive goal selection, and exploit the compositional nature of language~\cite{jiang2019hrl} to improve generalizability. 

We develop a two-level framework, \textbf{H}ierarchical \textbf{K}nowledge \textbf{Graph}-based \textbf{A}gent (H-KGA)\footnote{Code is available at: \url{https://github.com/YunqiuXu/H-KGA}}, to learn a hierarchy of policies with the aid of KG. 
In the high level, we use a meta-policy to obtain a set of available goals characterized by texts, and select one of them according to the current KG-based observation.
Then, we use a sub-policy for goal-conditioned reinforcement learning. 
Besides, we design a scheduled training strategy to facilitate learning across multiple levels. 
We conduct experiments on a series of cooking games across 8 levels, while only 4 levels are available during training.
The experimental results show that our method improves generalizability in both seen and unseen levels. 

Our contributions are summarised as follows:
Firstly, we are the first to study generalizability in text-based games from the aspect of hierarchical reinforcement learning.
Secondly, we develop a two-level HRL framework leveraging the KG-based observation for adaptive goal selection and goal-conditioned decision making. 
Thirdly, we empirically validate the effectiveness of our method in games with both seen and unseen difficulty levels, which show favorable generalizability.

\section{Related work}
\subsection{RL agent for text-based games} 
Motivated by the prosperity of deep reinforcement learning techniques in playing games~\cite{silver2016mastering}, robotics~\cite{schulman2017ppo,fang2019dher,fang2019curriculum} and NLP~\cite{fang2017learning}, several RL-based game agents have been developed for text-based games~\cite{he2016drrn,yuan2018counting,jain2019aaai,yin2019cog,guo2020emnlp,xu2020cog}. Compared with the non-learning-based agents ~\cite{hausknecht2019nail, atkinson2019text}, the RL-based agents are more favorable as there is no need to handcraft game playing strategies with huge amounts of expert knowledge. The KG-based agents~\cite{murugesan2020kg,xu2020nips} extend RL-based agents with the knowledge graph, which can be constructed from the raw textual observation via simple rules~\cite{ammanabrolu2019kgdqn}, language models~\cite{ammanabrolu2020kga2cexplore} or pre-training tasks~\cite{adhikari2020gatav2}. The major benefit of KG is that it serves as a belief state to provide structural and historical information to handle partial observability. While these works focus on constructing KG from the textual observation, we aim at improving generalizability by fully exploiting the KG to design a goal-conditioned HRL. Our work thus complements KG-based agents. 

\subsection{Generalization in text-based games} 
It may be difficult to study generalization in games initially designed for human players~\cite{hausknecht2019jericho}, as they are so challenging that existing RL agents are still far from being able to solve a large proportion of them even under the single game setting~\cite{yao2020emnlp}. Furthermore, these games usually have different themes, vocabularies and logics, making it hard to determine the domain gap~\cite{ammanabrolu2019kgdqntransfer}. Compared with these man-made games, the synthetic games~\cite{cote2018textworld,urbanek2019light} provide a more natural way to study generalization by generating multiple similar games with customizable domain gaps (e.g., by varying game layouts). Generally, the training and testing game sets in previous works have either the same difficulty level~\cite{ammanabrolu2019kgdqn,murugesan2020commonsense}, or a mixture of multiple levels~\cite{adolphs2019ledeepchef,yin2020emnlpfinding}, or both~\cite{adhikari2020gatav2}. 
In this work, we extend the setting of multiple levels to unseen levels. 
We not only study generalization in games that have the same difficulty level but various layouts, but also consider games where both the layouts and levels are different from those of the training games. 
In addition, we emphasize on improving the performance on hard levels. 

\begin{figure*}[t!]
\centering
\includegraphics[width=0.98\textwidth]{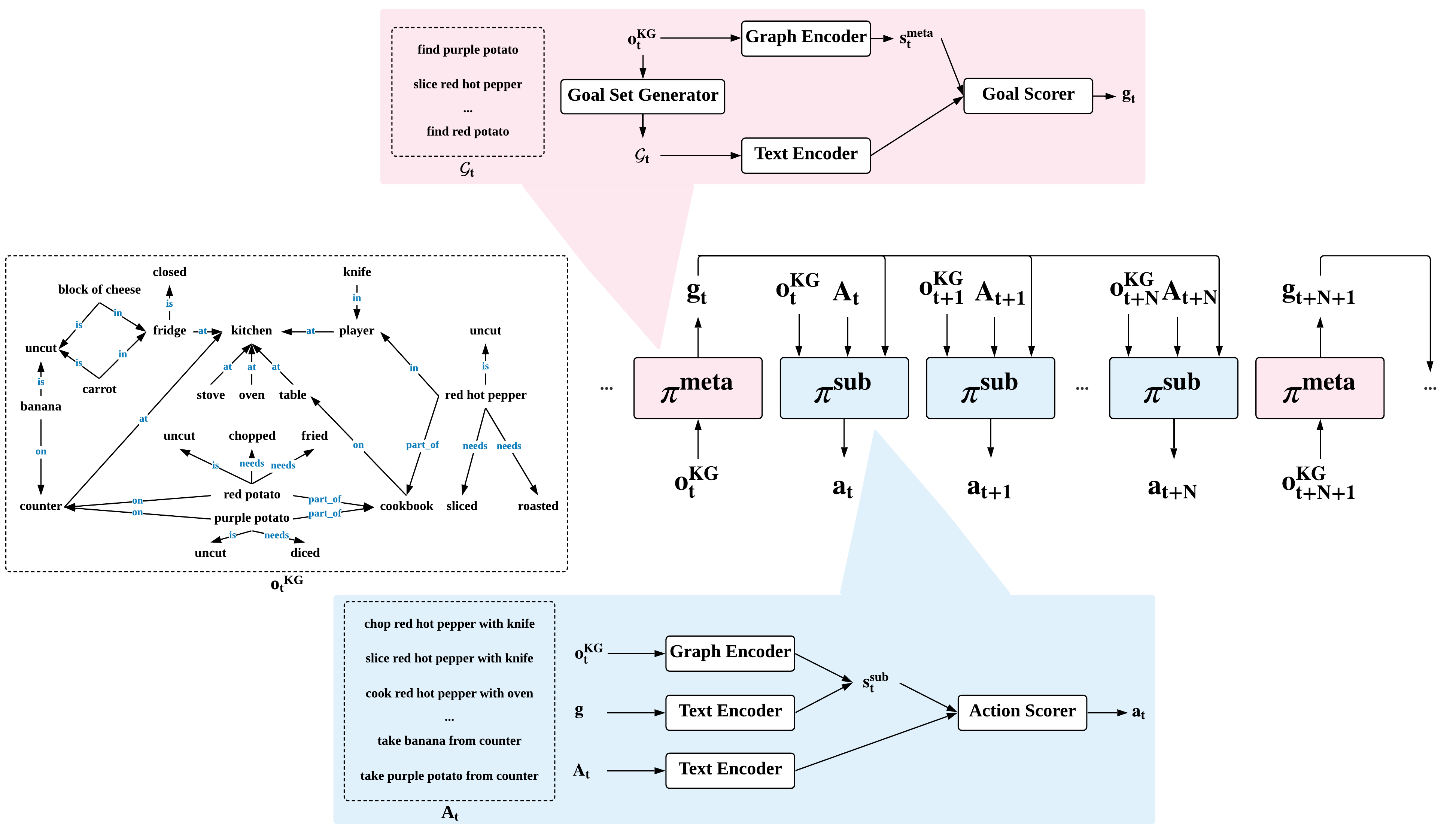}
\caption{The overview of H-KGA. In the high level (red), the meta-policy $\pi^{\text{meta}}$ first obtains the set of goals of  available subtasks $\mathcal{G}_t$ from $o_t^{\text{KG}}$, then selects a goal $g_t$ for the sub-policy $\pi^{\text{sub}}$.
In the low level (blue), $\pi^{\text{sub}}$ selects the action $a_t$ from the admissible action set $A_t$ conditioned on $o_t^{\text{KG}}$ and $g$.}
\label{emnlp_arch}
\end{figure*}

\subsection{Hierarchical reinforcement learning}
The HRL framework~\cite{dayan1992hrl_feudal} has been studied in video games~\cite{kulkarni2016hrl, vezhnevets2017hrl_feudal, shu2018hrl}, robotic control tasks~\cite{nachum2018hrl},and NLP tasks such as the dialogue system~\cite{peng2017HRLforDialogue,saleh2020HRLforDialogue}. 
However, as far as we know, we are the first to introduce the insight into text-based games with KG-based observation.  
Previous works also considered identifying a task by textual goal specifications~\cite{bahdanau2018goal_language,fu2018goal_language}.
In the domain of text-based games, such goal-conditioned RL setting has been studied with the quest generation tasks~\cite{ammanabrolu2019quest,ammanabrolu2020quest}. In our work, we specify a subtask by its goal. 
Different from these works, where a single goal is pre-specified or directly generated from the observation, we introduce a hierarchy by disentangling the process of goal set generation and goal selection. 
By accommodating flexible goal set generation (e.g., by pre-trained language models or human experts), we focus on designing a meta-policy to select the goal in an adaptive manner.  
By adopting HRL to select a textual-based goal for the sub-policy, our work is similar to HIN~\cite{jiang2019hrl} which, however, focuses on visual scenarios and separately trains the meta-policy and the sub-policy, leaving joint training as future work. 
Instead, we consider the domain of text-based games, and develop a framework to enable joint training of meta-policy and sub-policy. 
We further compare joint and individual training in Sec.  \ref{section_results_and_discussions}. 


\section{Background}

\subsection{KG-based observation}

Following previous works~\cite{hausknecht2019jericho}, we formulate the text-based games as Partially Observable Markov Decision Processes (POMDPs), where the details is in Appendix \ref{appendix_pomdp}. 
We discard the raw textual observation and consider only the KG-based observation $o_t^{\text{KG}}$  as the observational input at timestep $t$. 
Fig. \ref{emnlp_arch} shows an example of $o_t^{\text{KG}}$.
The KG is defined as $G=(V, E)$, where $V$ and $E$ are the node set and the edge set, respectively.  
$o_t^{\text{KG}}$ consists of a set of triplets, where a triplet is formulated as $\langle\mathit{Subject}$, $\mathit{Relation}$, $\mathit{Object}\rangle$, denoting that the $\mathit{Subject} \in V$ has $\mathit{Relation} \in E$ with the $\mathit{Object} \in V$.

\subsection{Problem setting}
We aim to design an RL-based agent that is able to address the generalization in solving text-based games. 
To reduce the requirement for external knowledge, we consider games sharing similar themes and vocabularies, but varying in their layouts and / or difficulty levels. 
For example, games of the cooking theme~\cite{cote2018textworld} share the same overall objective: prepare the meal.
To accomplish it, the player has to collect ingredients and prepare them in correct ways.
The layout of a game contains the room connectivity and the preparing steps (e.g., the type / location of ingredients).
The difficulty of a game depends on the complexity of the map (e.g., the number of rooms) and the recipe (e.g., the number of ingredients), such that two games with different levels are naturally different in their layouts. 
We follow the multi-task learning setting to consider that the training set and the testing set consist of multiple games from multiple levels. 
We consider two scenarios of generalization: 1) seen levels, where the training and testing games have the same levels, but different layouts. 2) unseen levels, where the training and testing games are different in levels and layouts. 
More examples are provided in Appendix. \ref{appendix_more_game_examples}.


\section{Methodology}

\subsection{Overview}

Fig. \ref{emnlp_arch} shows the overview of H-KGA, which consists of a hierarchy of two levels of policies.  
In the high level, a meta-policy $\pi^{\text{meta}}$ first obtains the set of goals of  available subtasks $\mathcal{G}_t$ from $o_t^{\text{KG}}$, then selects a goal $g_t$ for the sub-policy $\pi^{\text{sub}}$.
In the low level, $\pi^{\text{sub}}$ selects the action $a_t$ from the admissible action set $A_t$ conditioned on $o_t^{\text{KG}}$ and $g$. We omit the subscript ``t'' for $g$ because this goal may be selected in the past rather than the current time step. For example, as shown in Fig. \ref{emnlp_arch}, once the $g_t$ is selected by $\pi^{\text{meta}}$, it will remain unchanged for $N$ time steps until being completed (e.g., accomplished or failed). At each time step from $t$ to $t+N$, $\pi^{\text{sub}}$ considers the same goal $g_t$. 

In the following, we illustrate how to design $\pi^{\text{meta}}$ to obtain the available goal set $\mathcal{G}_t$ and conduct goal selection to obtain $g_t \in \mathcal{G}_t$ in Sec. \ref{section_method_part1}; how to design $\pi^{\text{sub}}$ to select an action $a_t \in A_t$ in Sec. \ref{section_method_part2}; 
and how to train H-KGA with a scheduled curriculum for multi-task learning in Sec. \ref{section_method_part3}. 

\subsection{Meta-policy for goal selection \label{section_method_part1}}

As discussed before, while a whole game may be difficult to accomplish due to long-term temporal dependency, decomposing it into a sketch of subtasks will make the game easier to be solved~\cite{sohn2018subtask,shiarlis2018taco}. 
If we consider the solving strategy for a subtask as a skill, the generalizability for an unseen game will also be improved by recomposing the learnt skills. 
Therefore, inspired by the HRL framework~\cite{sutton1999hrl}, we design a meta-policy $\pi^{\text{meta}}$ to first obtain a set of subtasks, then select one subtask from them. We characterize a subtask by its goal to transform subtask selection into goal selection.
We make the goal to be instruction-like textual descriptions (e.g., ``find purple potato''), yielding better flexibility and interpretability than using a state as the goal~\cite{andrychowicz2017her}.
Fig. \ref{emnlp_arch} shows the overview of $\pi^{\text{meta}}$ (in red), which consists of a goal set generator, a graph encoder, a text encoder and a goal scorer. We denote the set containing all required goals for solving a game as $\mathcal{G}$.
Then we define a goal as ``available'' at a time step if no other goals should be accomplished before it. For example, ``cook red potato'' is not available in Fig. \ref{emnlp_arch}, as another goal ``find red potato'' should be accomplished first. The goal set generator has two purposes: 1) obtain the set of currently available goals $\mathcal{G}_t \subseteq \mathcal{G}$, and 2) check whether a goal has been accomplished. 
Inspired by~\cite{jiang2019hrl}, the goal set generator can be implemented by different approaches, including supervised language models and non-learning-based methods such as human supervisors and functional programs. 
In our work, we use a non-learning-based method to obtain $\mathcal{G}_t$ and the details are discussed in Sec~\ref{section_implementation} and Appendix \ref{appendix_goal_set_generator}.

After obtaining $\mathcal{G}_t$, $\pi^{\text{meta}}$ will be used to select a goal $g_t \in \mathcal{G}_t$. 
We use a graph encoder to encode $o_t^{\text{KG}}$ as state representation $\bm{s}_t^{\text{meta}}$, and a text encoder to encode $\mathcal{G}_t$ as a stack of goal representations. 
Arbitrary graph encoders and text encoders can be used. 
We implement the graph encoder based on the Relational Graph Convolutional Networks (R-GCNs)~\cite{schlichtkrull2018rgcn} to take both nodes and edges into consideration. 
For the text encoder, a simple single-block transformer ~\cite{vaswani2017transformer} is sufficient as the goal candidates are short texts. 
In the goal scorer, we adopt a goal scoring process similar to~\cite{he2016drrn}, where $\bm{s}_t^{\text{meta}}$ will be paired with each goal representation, then processed by linear layers to obtain the goal scores. 
The scores can be treated as either sampling probabilities or Q values, where the goal candidate with the highest Q value will be selected. 

Following the Semi-Markov Decision Process (SMDPs)~\cite{sutton1999hrl}, $\pi^{\text{meta}}$ will be re-executed once a goal is accomplished / failed. $\pi^{\text{meta}}$ receives rewards $r_t^{\text{env}}$ from the environment. 
In a transition for $\pi^{\text{meta}}$, the reward is set as the sum of environment rewards:
\begin{equation}
    r^{\text{meta}} = \sum_{i=1}^T r^{\text{env}}_{t+i}
\end{equation}
where $T$ denotes time steps for accomplishing $g_t$. 

\subsection{Sub-policy for action selection \label{section_method_part2}}

The sub-policy $\pi^{\text{sub}}$ follows the goal-conditioned RL setting~\cite{kaelbling1993goal} where $a_t$ is selected by considering both $o_t^{\text{KG}}$ and $g$.
Fig. \ref{emnlp_arch} shows the architecture of $\pi^{\text{sub}}$ (in blue), which is similar to $\pi^{\text{meta}}$ except that the state $\bm{s}_t^{\text{sub}}$ is constructed based on both $o_t^{\text{KG}}$ and $g$. 
The graph encoder and text encoder in $\pi^{\text{meta}}$ can be re-used in $\pi^{\text{sub}}$, or be re-initialized with new weights. 
As this work does not aim at handling the combinatorial action space, we consider the admissible action set $A_t \subseteq \mathcal{A}$ for each time step. We denote an action as ``admissible'' if it does not lead to meaningless feedback (e.g., ``Nothing happens'').
Similar to the goal scorer in $\pi^{\text{meta}}$,  the action scorer will pair  $\bm{s}_t^{\text{sub}}$ with each candidate $a_i \in A_t$, followed by linear layers to compute the action scores. 

Depending on goal accomplishment, $\pi^{\text{sub}}$ receives binary intrinsic reward $r^{\text{goal}}_t \in \{ r_{\text{min}}, r_{\text{max}}\}$, which in this work can be determined by reusing the goal set generator upon $o_{t+1}^{\text{KG}}$.
Take Fig. \ref{emnlp_arch} as an example. If the goal before observing $o_t^{\text{KG}}$ is ``find knife'', the agent will receive $r^{\text{goal}}_t = r_{\text{max}}$, as this goal is accomplished at time step $t$.
Although the KG can serve as a ``map'' to provide guidance, such binary reward is insufficient for the agent to accomplish  a goal in complex games (e.g., the agent has to go through multiple rooms to find an ingredient).
To further improve the performance of $\pi^{\text{sub}}$, we reshape
the sub-reward with the count-based intrinsic reward~\cite{bellemare2016counting} to encourage exploration. 
Specifically, we apply the BeBold method~\cite{zhang2020counting} to the text-based games domain. 
During training, we count the visitation of observations within an episode, and the accumulated visitation throughout the training process.
The count-based reward $r^{\text{count}}_t$ is then defined as the regulated difference of inverse cumulative visitation counts with episodic restriction:
\begin{equation}
\begin{split}
     r^{\text{count}}_{t+1} = \max ( & 
     \frac{1}{\mathrm{N}_{\text{acc}}(o_{t}^{\text{KG}})} - \frac{1}{\mathrm{N}_{\text{acc}}(o_{t+1}^{\text{KG}})}, 
     \\ & 0) \cdot \mathbb{I}\{\mathrm{N}_{\text{epi}}(o_{t+1}^{\text{KG}}) = 1\}   
\end{split}
\label{eq_bebold}
\end{equation}
where $\mathrm{N}_{\text{acc}}$ and $\mathrm{N}_{\text{epi}}$ denote the accumulated and episodic visitation count, respectively.
The $\mathbb{I}$ operation returns 1 if $o_{t+1}^{\text{KG}}$ is visited for the first time in the current episode, otherwise 0. 
The reward for $\pi^{\text{sub}}$ can then be obtained by combining $r^{\text{goal}}_{t+1}$ and $r^{\text{count}}_{t+1}$: 
\begin{equation}
    r^{\text{sub}}_{t+1} = r^{\text{goal}}_{t+1} + \lambda \cdot r^{\text{count}}_{t+1}
\label{eq_sub}
\end{equation}
where $\lambda$ is a constant coefficient. 

\begin{algorithm}[t!]
\small
\caption{Training Strategy for H-KGA}
\label{algo_Training}
\textbf{Input:} 
game sets $\{ \mathcal{D}_{\text{train}}, \mathcal{D}_{\text{val}} \}$,
replay buffers $\{B^{\text{meta}}, B^{\text{sub}}\}$, 
update frequencies $\{F_{\text{up}}^{\text{meta}},F_{\text{up}}^{\text{sub}}\}$,
validation frequency $F_{\text{val}}$,
tolerance $\tau$,
coefficients $\beta$, $\lambda$, 
patience $P$ \\
\textbf{Initialize:} 
counters $k \leftarrow 1$, $p \leftarrow 0$, 
$N_{\text{acc}} \leftarrow \emptyset$, $N_{\text{epi}} \leftarrow \emptyset$, 
best validation score $V_{\text{val}} \leftarrow 0$, 
$r^{\text{meta}} \leftarrow 0$,
caches $\{ C^{\text{meta}}, C^{\text{sub}} \}$, 
policies $\{ \pi^{\text{meta}}, \pi^{\text{sub}} \}$, $\{ \Pi^{\text{meta}}, \Pi^{\text{sub}} \}$
\begin{algorithmic}[1]
\For{$e \leftarrow 1$ to $\text{NUM\_EPISODES}$}
    \State $l \leftarrow \text{SampleLevel}(\mathcal{L },p_l)$
    \State $x \leftarrow \text{SampleGame}(\mathcal{D}_{\text{train}}, l)$
    \State $o_0^{\text{KG}} \leftarrow$ reset $x$ 
    \State $C^{\text{meta}} \leftarrow \emptyset$,
           $C^{\text{sub}} \leftarrow \emptyset$, 
           $N_{\text{epi}} \leftarrow \emptyset$,
    \State Update $N_{\text{acc}}$, $N_{\text{epi}}$ with $o_0^{\text{KG}}$
    \For{$t \leftarrow 0$ to $\text{NUM\_STEPS}$}
        \State $g \leftarrow \pi^{\text{meta}}(g|o_t^{\text{KG}})$
        \State $r^{\text{meta}} \leftarrow 0$
        \While {$g$ is not terminated}
            \State $a_t \leftarrow \pi^{\text{sub}}(a|o_t^{\text{KG}}, g)$
            \State Execute $a_t$, receive $o_{t+1}^{\text{KG}}$, $r_{t+1}^{\text{env}}$, obtain $r_{t+1}^{\text{goal}}$
            \State Update $N_{\text{acc}}$, $N_{\text{epi}}$ with $o_{t+1}^{\text{KG}}$
            \State Compute $r_{t+1}^{\text{sub}}$ using Eq. (\ref{eq_bebold}) and Eq. (\ref{eq_sub})
            \State Store the sub transition into $C^{\text{sub}}$
            \State $r^{\text{meta}} \leftarrow r^{\text{meta}} + r_{t+1}^{\text{env}}$
            \State $t \leftarrow t+1$
            \State $k \leftarrow k+1$
            \If {$k \% F_{\text{up}}^{\text{meta}} = 0$} 
                \State $\text{Update}(\pi^{\text{meta}}, B^{\text{meta}})$
            \EndIf
            \If {$k \% F_{\text{up}}^{\text{sub}} = 0$} 
                \State $\text{Update}(\pi^{\text{sub}}, B^{\text{sub}})$
            \EndIf
        \EndWhile
        \State Store the meta transition into $C^{\text{meta}}$
    \EndFor
    \State Update $p_l$ using Eq. (\ref{eq_pl})
    \If {$\text{Avg}(r^{\text{meta}}|C^{\text{meta}}, l) > \tau \cdot \text{Avg}(r^{\text{meta}}|B^{\text{meta}}, l)$}
        \State Store all transitions in $C^{\text{meta}}$ into $B^{\text{meta}}$
    \EndIf
    \If {$\text{Avg}(r^{\text{goal}}|C^{\text{sub}}, l) > \tau \cdot \text{Avg}(r^{\text{goal}}|B^{\text{sub}}, l)$}
        \State Store all transitions in $C^{\text{sub}}$ into $B^{\text{sub}}$
    \EndIf
    \If {$e \% F_{\text{val}} = 0$}
        \State $v_{\text{val}} \leftarrow \text{Validate}(\pi^{\text{meta}},\pi^{\text{sub}}, \mathcal{D}_{\text{val}})$
        \If {$v_{\text{val}} \ge V_{\text{val}}$}
            \State $V_{\text{val}} \leftarrow v_{\text{val}}$, $\Pi^{\text{meta}} \leftarrow \pi^{\text{meta}}$, $\Pi^{\text{sub}} \leftarrow \pi^{\text{sub}}$
            \State $p \leftarrow 0$, continue
        \EndIf
        \If {$p > P$}
            \State $\pi^{\text{meta}} \leftarrow \Pi^{\text{meta}}$, $\pi^{\text{sub}} \leftarrow \Pi^{\text{sub}}$, $p \leftarrow 0$
        \Else 
            \State $p \leftarrow p+1$
        \EndIf
    \EndIf
\EndFor
\end{algorithmic}
\end{algorithm}

\subsection{Training H-KGA for multi-task learning \label{section_method_part3}}

We train H-KGA via Double DQN~\cite{hasselt2016doubledqn} with prioritized experience replay~\cite{schaul2015per}. 
Algo. \ref{algo_Training} shows the training strategy. 
We consider a training set $\mathcal{D}_{\text{train}}$ with $\mathcal{L}$ levels of games. 
For each episode, we sample a game $x$ from $\mathcal{D}_{\text{train}}$ to interact with (lines 2-22). 
A goal $g$ will be terminated if it is accomplished$/$ failed, or $t$ exceeds $\text{NUM\_STEPS}$. 
We formulate the meta transition as 
         $\langle o_{t}^{\text{KG}}, 
         g, 
         r^\text{meta}, 
         o_{t+T}^{\text{KG}}, l \rangle$, 
and the sub transition as 
         $\langle (o_{t}^{\text{KG}}, g), 
         a_t, 
         r_{t+1}^{\text{sub}}, r_{t+1}^{\text{goal}}, 
         (o_{t+1}^{\text{KG}}, g), l \rangle$, where $l \in \mathcal{L}$ denotes the level of a game.
We update $\pi^{\text{meta}}$ ($\pi^{\text{sub}}$) per $F_{\text{up}}^{\text{meta}}$ ($F_{\text{up}}^{\text{sub}}$) interaction steps, by sampling a batch of transitions from the replay buffer $B^{\text{meta}}$ ($B^{\text{sub}}$). 
In addition, we leverage two strategies empirically effective for previous agents~\cite{adhikari2020gatav2}. 
First, we collect the episodic transitions within a cache, and only push them into the replay buffer when its average reward is greater than $\tau$ times the average reward of the buffer (lines 23-26).
Second, we validate the model on a validation set $\mathcal{D}_{\text{val}}$ per $F_{\text{val}}$ episodes and keep track of the best score $V$ and the corresponding policies $\{ \Pi^{\text{meta}}, \Pi^{\text{sub}} \}$. We load the training policies $\{ \pi^{\text{meta}}, \pi^{\text{sub}} \}$ back to $\{ \Pi^{\text{meta}}, \Pi^{\text{sub}} \}$, if the validation performance $v_{\text{val}}$ keeps being worse than $V$ for over $P$ times (lines 27-35). 

The training process can be formulated as multi-task learning if we treat learning on games from the same level as a task. 
While the knowledge can be shared across levels, different levels may have different scales of training time and performance.
For example, those from hard levels generally require more time to learn and tend to have lower normalized performance. 
To facilitate such multi-task learning setting, we further propose two strategies to improve Algorithm \ref{algo_Training}: 1) scheduled task sampling and 2) level-aware replay buffer. 
The scheduled task sampling is  inspired by the curriculum learning~\cite{bengio2009curriculum}, where we schedule the tasks based on their difficulties. 
We track the training performance $v_l$ on a level $l$, and compute the sampling probability as: 
\begin{equation}
    p_{l} = \frac{\text{exp}(\beta - v_{l})}{\sum_{l_i \in \mathcal{L}} \text{exp}(\beta - v_{l_i})}
\label{eq_pl}
\end{equation}
where $\beta$ is a constant coefficient. 
For each episode, we first sample a level based on the probabilities, and then sample a training game from this level uniformly (lines 2-3). 
Compared to level-invariant sampling, this strategy encourages the agent to focus more on hard levels with training going on. 
Another strategy, level-aware replay buffer, is conducted when moving transitions from cache to the replay buffer (lines 23-26). 
As the transitions collected from hard games tend to have lower reward, they are not likely to be added to the replay buffer. 
To alleviate this problem, we make the level as an additional component of transition and record the average reward of each level.
Then we compare those belonging to the same level to determine whether to add new transitions. 

\section{Experiments}
\subsection{Experiment setting}

We conduct experiments on multiple levels of cooking games~\cite{cote2018textworld}. 
While previous work~\cite{adhikari2020gatav2} considered either a single level, or a mixture of 4 levels, we extend their setting to 8 levels. 
Based on the rl.0.1 game set\footnote{\url{https://aka.ms/twkg/rl.0.1.zip}}, we build a training game set $\mathcal{D}_{\text{train}}$ with 4 levels, including 100 games per level. 
We build a validating game set $\mathcal{D}_{\text{val}}$ with the same 4 levels of $\mathcal{D}_{\text{train}}$, where each level contains 20 games.
We build two testing game sets: $\mathcal{D}_{\text{test}}^{\text{seen}}$, and $\mathcal{D}_{\text{test}}^{\text{unseen}}$, both of which contain 4 levels and 20 games per level.
The levels within $\mathcal{D}_{\text{test}}^{\text{seen}}$ have been seen in $\mathcal{D}_{\text{train}}$ and $\mathcal{D}_{\text{val}}$, while there is no overlapping game. 
The levels within $\mathcal{D}_{\text{test}}^{\text{unseen}}$ are unseen during training. 
Table \ref{table_game_statistics} shows the game statistics averaged over each level, where ``S\#'' denotes a seen level and ``US\#'' denotes an unseen level.

\begin{table}[t!]
\caption{\label{table_game_statistics} Game statistics. ``\#Ings'' denotes the number of ingredients, ``\#Reqs'' denotes the requirements, and ``\#Acts'' denotes the admissible actions per time step.}
\centering
\resizebox{0.49\textwidth}{!}{
\begin{tabular}{cccccccc}
\hline 
\textbf{Level}  & \textbf{\#Triplets} & \textbf{\#Rooms} & \textbf{\#Objs} & \textbf{\#Ings} & \textbf{\#Reqs} & \textbf{\#Acts} & \textbf{MaxScore} \\
\hline
S1  & 21.44 & 1  & 17.09 & 1 & 1 & 11.54 & 4 \\ 
S2  & 21.50 & 1  & 17.49 & 1 & 2 & 11.81 & 5 \\ 
S3  & 46.09 & 9  & 34.15 & 1 & 0 & 7.25  & 3 \\ 
S4  & 54.54 & 6  & 33.41 & 3 & 2 & 28.38 & 11 \\
\hline
US1 & 19.85 & 1  & 16.01 & 1 & 0 & 7.98 & 3 \\ 
US2 & 20.74 & 1  & 16.69 & 1 & 1 & 8.87 & 4 \\ 
US3 & 33.04 & 6  & 24.81 & 1 & 0 & 7.61 & 3 \\
US4 & 47.31 & 6  & 31.09 & 3 & 0 & 13.90& 5 \\
\hline
\end{tabular}
}
\end{table}

\subsection{Baselines}

We consider the following five models, and compare with more variants in ablation studies:
\begin{itemize}
    \item GATA~\cite{adhikari2020gatav2}: a powerful KG-based agent and the state-of-the-art on the rl.0.1 game set. However, it does not have hierarchical architecture, and the action selection policy is not goal-conditioned.
    \item GC-GATA: GATA equipped with a goal set generator, a \textbf{g}oal-\textbf{c}onditioned action selection (sub-)policy, and a non-learnable meta-policy for random goal selection.
    \item H-KGA: the proposed model with both meta-policy and sub-policy.
    \item H-KGA HalfJoint: an H-KGA variant, where during the first \textbf{half} of training process only the sub-policy is trained, then the two policies are \textbf{joint}ly trained. 
    \item H-KGA Ind: an H-KGA variant, where the two policies are \textbf{ind}ividually trained (the sub-policy for the first half, then the meta-policy). 
\end{itemize}

\subsection{Implementation details\label{section_implementation}}
We implement the models based on GATA's released code\footnote{\url{https://github.com/xingdi-eric-yuan/GATA-public}}.
In particular, we adopt the version GATA-GTF and denote it as GATA for simplicity. 
GATA-GTF discards textual observations and uses the ground truth full KG as observation, so that there is no information extraction error incurred during KG construction. 
We design a simple non-learning-based goal set generator to obtain available goals (leaving pre-training-based generators as future work). Please refer to Appendix \ref{appendix_goal_set_generator} for details.
All models follow the same architecture of graph encoder (i.e., R-GCNs), text encoder (i.e., single transformer block with single head) and scorers (i.e., linear layers). 
The encoders in $\pi^{\text{meta}}$ and $\pi^{\text{sub}}$ are initialized separately. 

We set the step limit of an episode as 50 for training and 100 for validation / testing. We train the models for 100,000 episodes. All models apply the BeBold reward bonus with $\lambda = 0.1$, and the scheduled sampling method with $\beta = 1.0$. 
We set $B^{\text{meta}}$ with size 50,000 and $B^{\text{sub}}$ with size 500,000.
We set $F_{\text{up}}^{\text{meta}}$ and $F_{\text{up}}^{\text{sub}}$ as 50 time steps, and the updating starts after 100 episodes with batch size 64.
The GC-GATA pre-trained for 50,000 episodes is used for initializing 
H-KGA HalfJoint and H-KGA Ind. 
For every 1,000 episodes, we validate the model on $\mathcal{D}_{\text{val}}$, and report the testing performance on $\mathcal{D}_{\text{test}}^{\text{seen}}$ and $\mathcal{D}_{\text{test}}^{\text{unseen}}$. 
The experiments are conducted on a Quadro RTX 6000 GPU. Each experiment is run with 3 random seeds, and each run takes 2-3 days to finish.

\subsection{Evaluation metrics}
We denote a game's score as the episodic sum of rewards without discount. 
We use the normalized score, which is defined as the collected score normalized by the maximum available score for this game, to measure the performance.
For each testing game set, we report the performance on each level and the performance averaged over levels. 

\section{Results and discussions \label{section_results_and_discussions}}

\begin{table}[t!]
\caption{\label{table_result_main} 
The testing performance at the end of training. }
\centering
\resizebox{0.49\textwidth}{!}{
\begin{tabular}{l|ccc}
\hline 
\textbf{Model}       & \textbf{Avg Seen} & \textbf{Avg Unseen} & \textbf{Avg All} \\
\hline
GATA                 & 0.47$\pm$0.04     & 0.62$\pm$0.07       & 0.55$\pm$0.06    \\
GC-GATA              & 0.54$\pm$0.13     & 0.61$\pm$0.12       & 0.58$\pm$0.12    \\
\textbf{H-KGA (ours)}         & \textbf{0.72}$\pm$0.04     & \textbf{0.79}$\pm$0.04       & \textbf{0.76}$\pm$0.03   \\
H-KGA HalfJoint      & 0.56$\pm$0.11     & 0.57$\pm$0.07       & 0.57$\pm$0.09    \\
H-KGA Ind            & 0.70$\pm$0.02     & 0.68$\pm$0.01       & 0.69$\pm$0.02    \\
\hline
GATA w/o BeBold      & 0.54$\pm$0.06     & 0.68$\pm$0.02       & 0.61$\pm$0.03    \\
H-KGA w/o BeBold     & 0.57$\pm$0.07     & 0.65$\pm$0.09       & 0.61$\pm$0.07    \\
\hline
H-KGA w/o Sch        & 0.52$\pm$0.07    & 0.63$\pm$0.07       & 0.57$\pm$0.05     \\
H-KGA w/o Sch w/o LR & 0.63$\pm$0.14     & 0.63$\pm$0.16       & 0.63$\pm$0.15    \\
\hline
\end{tabular}
}
\end{table}

\begin{figure}[t!]
\centering
\includegraphics[width=0.48\textwidth]{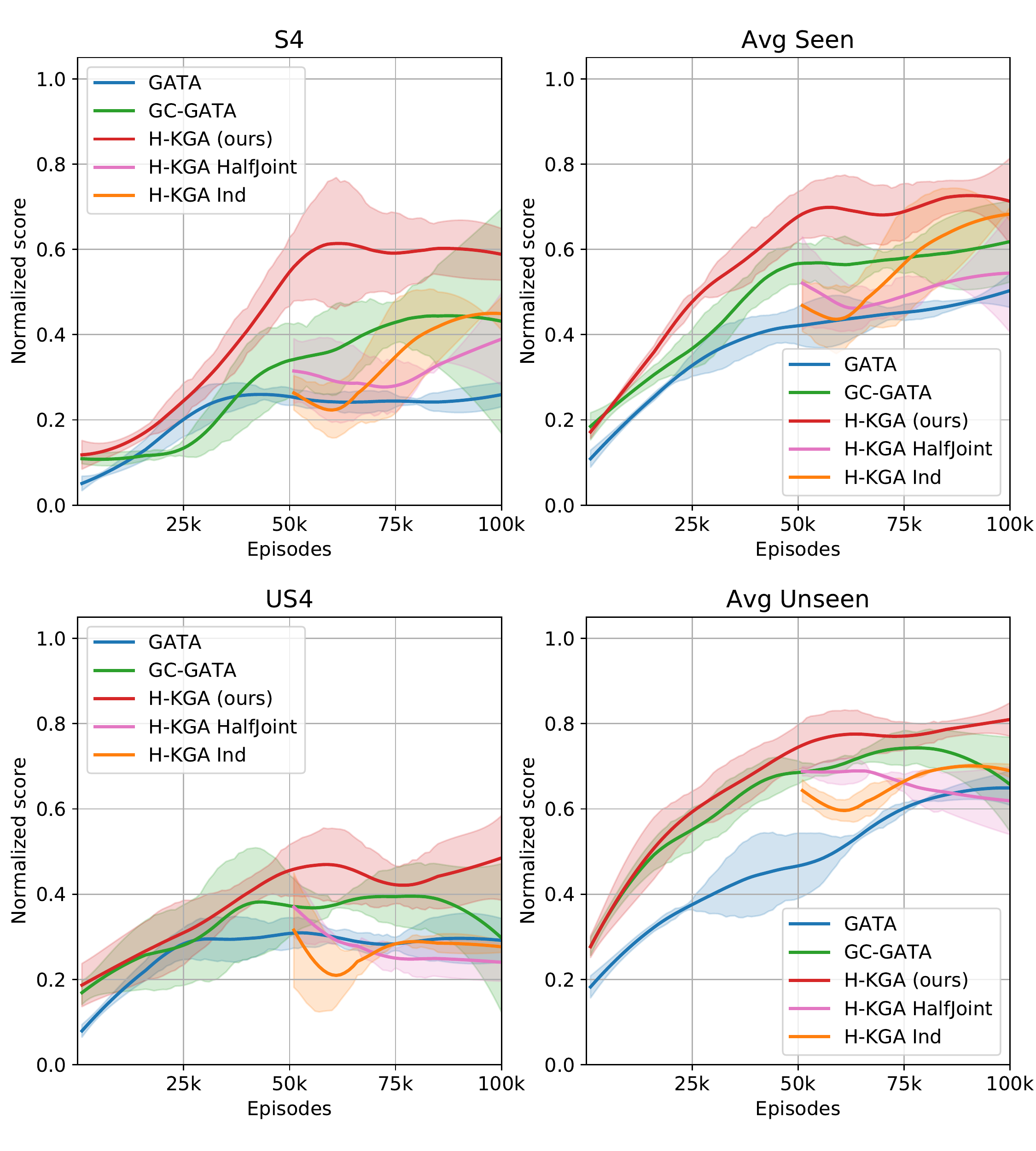}
\caption{The models' performance on $\mathcal{D}_{\text{test}}^{\text{seen}}$ (``S4'', ``Avg Seen'') and $\mathcal{D}_{\text{test}}^{\text{unseen}}$ (``US4'', ``Avg Unseen''). }
\label{exp1main_main}
\end{figure}

\subsection{Main results \label{section_exp_main_result}}

Table \ref{table_result_main} shows the testing performance at the end of training, and Fig. \ref{exp1main_main} shows the models' testing performance with respect to the training episodes. 
Due to space constraint, we present only results on the two most difficult levels, ``S4'' and ``US4'', as well as the average performance on $\mathcal{D}_{\text{test}}^{\text{seen}}$ and $\mathcal{D}_{\text{test}}^{\text{unseen}}$.
Please refer to Appendix \ref{appendix_more_exp_results} for the full results. 
Our H-KGA outperforms baselines in both seen and unseen levels. 
Its advantage becomes most significant in the most complex level, ``S4'', which is with the most number of rooms, ingredients and required preparation steps as shown in Table \ref{table_game_statistics}. 
The performance improvement of our model can be attributed to two aspects: the goal-conditioned sub-policy and the meta-policy for adaptive goal selection. 
GC-GATA, which can be regarded as H-KGA without the meta-policy, also achieves improvement over GATA, demonstrating the effectiveness of goal-conditioned decision making.
Compared to GC-GATA, the use of a learned meta-policy helps to further improve H-KGA.

However, we observe that joint training after pre-training the sub-policy leads to performance drop (H-KGA HalfJoint), which could be attributed to the forgetting problem in RL~\cite{vinyals2019alphastar}.
Another variant, H-KGA Ind, where the pre-trained sub-policy is frozen during training the meta-policy, performs better and exceeds GC-GATA, but still worse than our H-KGA.
While H-KGA Ind might still have space for performance improvement, it requires more training episodes (i.e., collecting more interaction samples), leading to low sample efficiency. 
Instead, our H-KGA utilizes the training data more efficiently and achieves comparable performance with fewer episodes, making it more favorable for practical applications. 

We also observe that learning a good meta-policy helps in solving games from unseen levels. 
In ``US4'', where the agent has to navigate through multiple rooms to collect three ingredients, it is more important to learn a strategy to determine the collecting order. 
In these games, our H-KGA performs better than those without a meta-policy (GATA, GC-GATA), and those with a "not-so-good" meta-policy (H-KGA HalfJoint, H-KGA Ind). 

\begin{figure}[t!]
\centering
\includegraphics[width=0.48\textwidth]{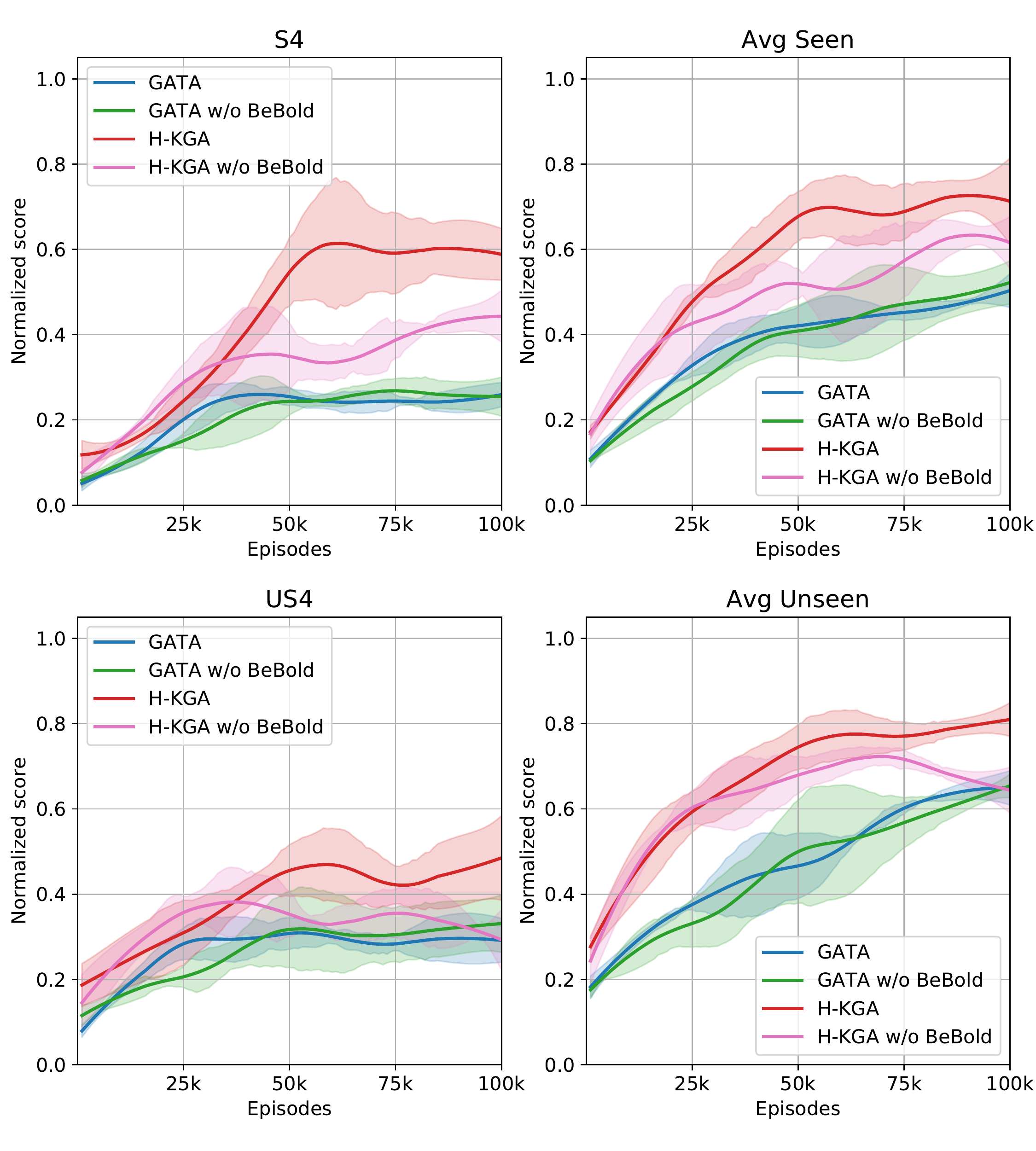}
\caption{The models' performance with / without the BeBold reward bonus.}
\label{exp2main_woBeBold}
\end{figure}

\subsection{The influence of exploration \label{section_exp_bebold}}

In Sec. \ref{section_method_part2}, we enhance the sub-policy with the BeBold reward to encourage exploration. 
We investigate its contribution by comparing models without such rewards. 
Fig. \ref{exp2main_woBeBold} shows the results. 
In terms of the average performance, our H-KGA is already better than GATA even without the BeBold reward (``H-KGA w/o BeBold'' v.s., ``GATA w/o BeBold''). 
However, the results on ``S4'' and ``US4'' show that sufficient exploration is essential for these difficult games, where it's hard for H-KGA without BeBold to collect over 50\% (40\%) of the scores in ``S4'' (``US4''). 
We also find that encouraging exploration only is not sufficient, as there is no obvious improvement for GATA, or even worse performance according to Table \ref{table_result_main}. 

\begin{figure}[t!]
\centering
\includegraphics[width=0.48\textwidth]{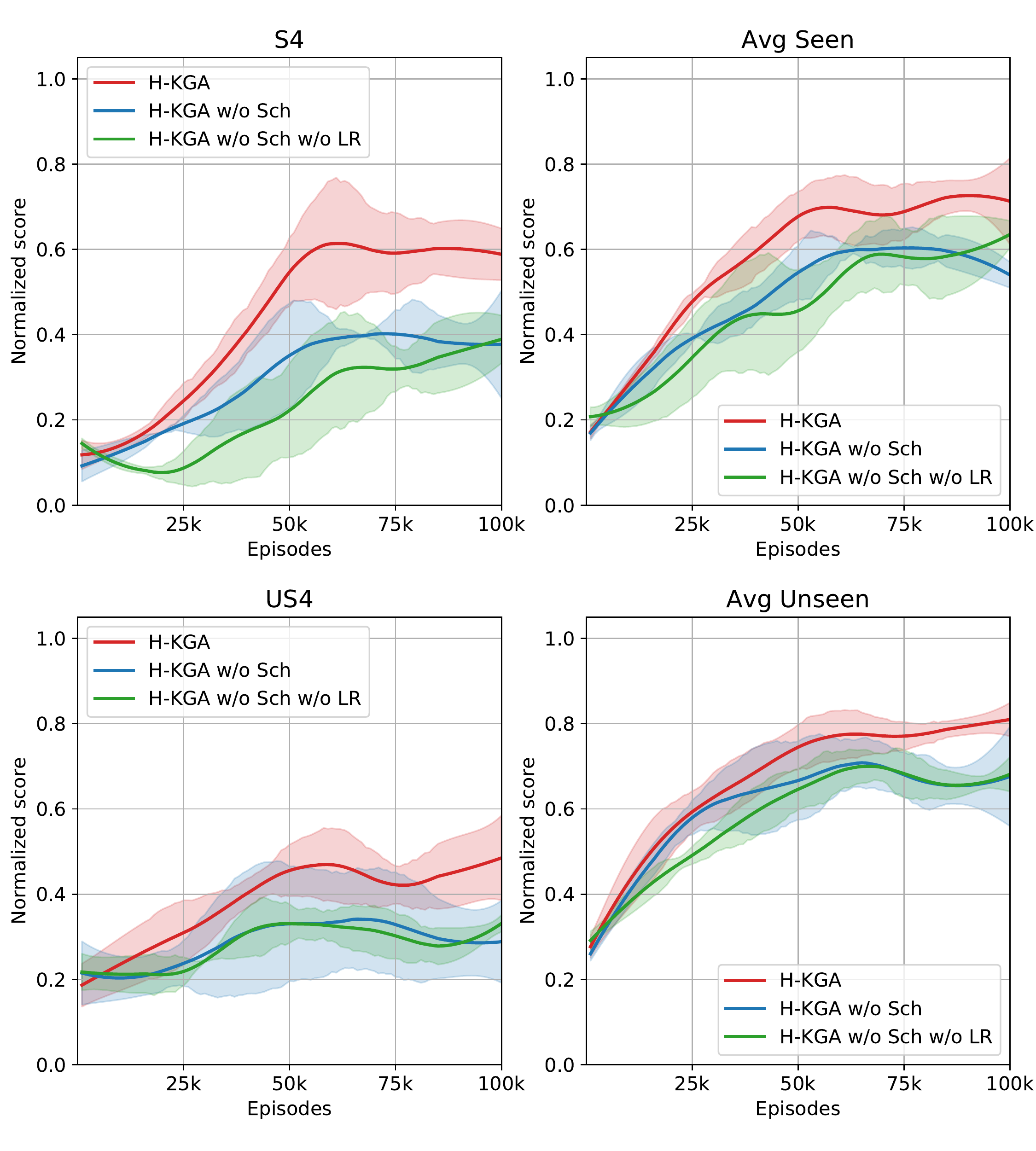}
\caption{The models' performance with / without the multi-task learning strategies. }
\label{exp3main_woMTL}
\end{figure}

\subsection{The influence of MTL strategies \label{section_exp_mtl}}

In Sec. \ref{section_method_part3}, we introduce two strategies to facilitate training H-KGA in the setting of multi-task learning.
We then conduct ablation studies to investigate their contributions.
Fig. \ref{exp3main_woMTL} shows the results, where ``Sch'' denotes the \textbf{sch}eduled task sampling and ``LR'' denotes \textbf{l}evel-aware \textbf{r}eplay buffer.
Although H-KGA can still achieve comparable average performance in both seen and unseen levels, without scheduled task sampling its performance on difficult levels, which require more training steps to collect more training samples, is limited.
Similarly, training without ``LR'' prevents transitions of difficult levels from being added to the replay buffer, leading to low sample efficiency.

\section{Conclusion}

In this paper, we investigated generalization for reinforcement learning in text-based games. 
We introduced a two-level hierarchical framework, H-KGA, to address this problem. 
In the high level, a meta-policy is executed to decompose the whole game as subtasks characterized by textual goals, and select a goal based on the knowledge graph-based observation. In the low level, a sub-policy is executed to select action conditioned on the goal. 
Experimental results showed that H-KGA achieved favorable performance on games with various difficulty levels. 
As an ongoing work, we would like to study automatic goal generation methods. We are also interested in extending our work to more complex scenarios .

\section*{Acknowledgement}
This work was supported in part by ARC DP180100966 and Facebook Research. We thank Xingdi Yuan and Marc-Alexandre C{\^o}t{\'e} from Microsoft Research, and anonymous reviewers for suggestions. 



\bibliographystyle{acl_natbib}
\bibliography{reference}

\begin{thebibliography}{58}
\expandafter\ifx\csname natexlab\endcsname\relax\def\natexlab#1{#1}\fi

\bibitem[{Adhikari et~al.(2020)Adhikari, Yuan, C{\^o}t{\'e}, Zelinka, Rondeau,
  Laroche, Poupart, Tang, Trischler, and Hamilton}]{adhikari2020gatav2}
Ashutosh Adhikari, Xingdi Yuan, Marc-Alexandre C{\^o}t{\'e}, Mikul{\'a}{\v{s}}
  Zelinka, Marc-Antoine Rondeau, Romain Laroche, Pascal Poupart, Jian Tang,
  Adam Trischler, and Will Hamilton. 2020.
\newblock Learning dynamic belief graphs to generalize on text-based games.
\newblock In \emph{Advances in Neural Information Processing Systems
  (NeurIPS)}, volume~33, pages 3045--3057.

\bibitem[{Adolphs and Hofmann(2020)}]{adolphs2019ledeepchef}
Leonard Adolphs and Thomas Hofmann. 2020.
\newblock Ledeepchef: Deep reinforcement learning agent for families of
  text-based games.
\newblock In \emph{Proceedings of the AAAI Conference on Artificial
  Intelligence (AAAI)}, volume~34, pages 7342--7349.

\bibitem[{Ammanabrolu et~al.(2019)Ammanabrolu, Broniec, Mueller, Paul, and
  Riedl}]{ammanabrolu2019quest}
Prithviraj Ammanabrolu, William Broniec, Alex Mueller, Jeremy Paul, and Mark~O
  Riedl. 2019.
\newblock \href {https://www.aclweb.org/anthology/2019.ccnlg-1.1} {Toward
  automated quest generation in text-adventure games}.
\newblock In \emph{Proceedings of the 4th Workshop on Computational Creativity
  in Language Generation}, pages 1--12.

\bibitem[{Ammanabrolu and Hausknecht(2020)}]{ammanabrolu2019kga2c}
Prithviraj Ammanabrolu and Matthew Hausknecht. 2020.
\newblock \href {https://openreview.net/forum?id=B1x6w0EtwH} {Graph constrained
  reinforcement learning for natural language action spaces}.
\newblock In \emph{International Conference on Learning Representations
  (ICLR)}.

\bibitem[{Ammanabrolu and Riedl(2019{\natexlab{a}})}]{ammanabrolu2019kgdqn}
Prithviraj Ammanabrolu and Mark Riedl. 2019{\natexlab{a}}.
\newblock \href {https://doi.org/10.18653/v1/N19-1358} {Playing text-adventure
  games with graph-based deep reinforcement learning}.
\newblock In \emph{Proceedings of the Conference of the North American Chapter
  of the Association for Computational Linguistics: Human Language Technologies
  (NAACL-HLT)}, volume~1, pages 3557--3565.

\bibitem[{Ammanabrolu and
  Riedl(2019{\natexlab{b}})}]{ammanabrolu2019kgdqntransfer}
Prithviraj Ammanabrolu and Mark Riedl. 2019{\natexlab{b}}.
\newblock \href {https://doi.org/10.18653/v1/D19-5301} {Transfer in deep
  reinforcement learning using knowledge graphs}.
\newblock In \emph{Proceedings of the Thirteenth Workshop on Graph-Based
  Methods for Natural Language Processing (TextGraphs-13)}, pages 1--10.

\bibitem[{Ammanabrolu et~al.(2020)Ammanabrolu, Tien, Luo, and
  Riedl}]{ammanabrolu2020kga2cexplore}
Prithviraj Ammanabrolu, Ethan Tien, Zhaochen Luo, and Mark~O Riedl. 2020.
\newblock How to avoid being eaten by a grue: Exploration strategies for
  text-adventure agents.
\newblock \emph{arXiv preprint arXiv:2002.08795}.

\bibitem[{Ammanabrolu et~al.(2021)Ammanabrolu, Urbanek, Li, Szlam,
  Rockt{\"a}schel, and Weston}]{ammanabrolu2020quest}
Prithviraj Ammanabrolu, Jack Urbanek, Margaret Li, Arthur Szlam, Tim
  Rockt{\"a}schel, and Jason Weston. 2021.
\newblock \href {https://doi.org/10.18653/v1/2021.naacl-main.64} {How to
  motivate your dragon: Teaching goal-driven agents to speak and act in fantasy
  worlds}.
\newblock In \emph{Proceedings of Conference of the North American Chapter of
  the Association for Computational Linguistics: Human Language Technologies
  (NAACL-HLT)}, pages 807--833.

\bibitem[{Andreas et~al.(2017)Andreas, Klein, and Levine}]{andreas2017modular}
Jacob Andreas, Dan Klein, and Sergey Levine. 2017.
\newblock \href {https://proceedings.mlr.press/v70/andreas17a.html} {Modular
  multitask reinforcement learning with policy sketches}.
\newblock In \emph{International Conference on Machine Learning (ICML)},
  volume~70, pages 166--175.

\bibitem[{Andrychowicz et~al.(2017)Andrychowicz, Wolski, Ray, Schneider, Fong,
  Welinder, McGrew, Tobin, Abbeel, and Zaremba}]{andrychowicz2017her}
Marcin Andrychowicz, Filip Wolski, Alex Ray, Jonas Schneider, Rachel Fong,
  Peter Welinder, Bob McGrew, Josh Tobin, Pieter Abbeel, and Wojciech Zaremba.
  2017.
\newblock Hindsight experience replay.
\newblock In \emph{Advances in Neural Information Processing Systems
  (NeurIPS)}, volume~30, pages 5048--5058.

\bibitem[{Atkinson et~al.(2019)Atkinson, Baier, Copplestone, Devlin, and
  Swan}]{atkinson2019text}
Timothy Atkinson, Hendrik Baier, Tara Copplestone, Sam Devlin, and Jerry Swan.
  2019.
\newblock \href {https://doi.org/10.1109/TG.2019.2896017} {The text-based
  adventure ai competition}.
\newblock \emph{IEEE Transactions on Games}, 11(3):260--266.

\bibitem[{Bahdanau et~al.(2019)Bahdanau, Hill, Leike, Hughes, Kohli, and
  Grefenstette}]{bahdanau2018goal_language}
Dzmitry Bahdanau, Felix Hill, Jan Leike, Edward Hughes, Pushmeet Kohli, and
  Edward Grefenstette. 2019.
\newblock \href {https://openreview.net/forum?id=H1xsSjC9Ym} {Learning to
  understand goal specifications by modelling reward}.
\newblock In \emph{International Conference on Learning Representations
  (ICLR)}.

\bibitem[{Bellemare et~al.(2016)Bellemare, Srinivasan, Ostrovski, Schaul,
  Saxton, and Munos}]{bellemare2016counting}
Marc Bellemare, Sriram Srinivasan, Georg Ostrovski, Tom Schaul, David Saxton,
  and Remi Munos. 2016.
\newblock Unifying count-based exploration and intrinsic motivation.
\newblock In \emph{Advances in Neural Information Processing Systems
  (NeurIPS)}, volume~29, pages 1471--1479.

\bibitem[{Bengio et~al.(2009)Bengio, Louradour, Collobert, and
  Weston}]{bengio2009curriculum}
Yoshua Bengio, J{\'e}r{\^o}me Louradour, Ronan Collobert, and Jason Weston.
  2009.
\newblock Curriculum learning.
\newblock In \emph{International Conference on Machine Learning (ICML)}, pages
  41--48.

\bibitem[{Cobbe et~al.(2019)Cobbe, Klimov, Hesse, Kim, and
  Schulman}]{cobbe2019rlgeneralization}
Karl Cobbe, Oleg Klimov, Chris Hesse, Taehoon Kim, and John Schulman. 2019.
\newblock Quantifying generalization in reinforcement learning.
\newblock In \emph{International Conference on Machine Learning (ICML)},
  volume~97, pages 1282--1289.

\bibitem[{C\^ot\'e et~al.(2018)C\^ot\'e, K\'ad\'ar, Yuan, Kybartas, Barnes,
  Fine, Moore, Tao, Hausknecht, Asri, Adada, Tay, and
  Trischler}]{cote2018textworld}
Marc-Alexandre C\^ot\'e, \'Akos K\'ad\'ar, Xingdi Yuan, Ben Kybartas, Tavian
  Barnes, Emery Fine, James Moore, Ruo~Yu Tao, Matthew Hausknecht, Layla~El
  Asri, Mahmoud Adada, Wendy Tay, and Adam Trischler. 2018.
\newblock Textworld: A learning environment for text-based games.
\newblock \emph{arXiv preprint arXiv:1806.11532}.

\bibitem[{Dayan and Hinton(1992)}]{dayan1992hrl_feudal}
Peter Dayan and Geoffrey~E Hinton. 1992.
\newblock Feudal reinforcement learning.
\newblock In \emph{Advances in Neural Information Processing Systems
  (NeurIPS)}, volume~5.

\bibitem[{Fang et~al.(2017)Fang, Li, and Cohn}]{fang2017learning}
Meng Fang, Yuan Li, and Trevor Cohn. 2017.
\newblock \href {https://doi.org/10.18653/v1/D17-1063} {Learning how to active
  learn: A deep reinforcement learning approach}.
\newblock In \emph{Proceedings of the Conference on Empirical Methods in
  Natural Language Processing (EMNLP)}, pages 595--605.

\bibitem[{Fang et~al.(2019{\natexlab{a}})Fang, Zhou, Shi, Gong, Xu, and
  Zhang}]{fang2019dher}
Meng Fang, Cheng Zhou, Bei Shi, Boqing Gong, Jia Xu, and Tong Zhang.
  2019{\natexlab{a}}.
\newblock \href {https://openreview.net/forum?id=Byf5-30qFX} {{DHER}: Hindsight
  experience replay for dynamic goals}.
\newblock In \emph{International Conference on Learning Representations
  (ICLR)}.

\bibitem[{Fang et~al.(2019{\natexlab{b}})Fang, Zhou, Du, Han, and
  Zhang}]{fang2019curriculum}
Meng Fang, Tianyi Zhou, Yali Du, Lei Han, and Zhengyou Zhang.
  2019{\natexlab{b}}.
\newblock Curriculum-guided hindsight experience replay.
\newblock In \emph{Advances in Neural Information Processing Systems
  (NeurIPS)}, volume~32, pages 12602--12613.

\bibitem[{Fu et~al.(2019)Fu, Korattikara, Levine, and
  Guadarrama}]{fu2018goal_language}
Justin Fu, Anoop Korattikara, Sergey Levine, and Sergio Guadarrama. 2019.
\newblock \href {https://openreview.net/forum?id=r1lq1hRqYQ} {From language to
  goals: Inverse reinforcement learning for vision-based instruction
  following}.
\newblock In \emph{International Conference on Learning Representations
  (ICLR)}.

\bibitem[{Guo et~al.(2020)Guo, Yu, Gao, Gan, Campbell, and
  Chang}]{guo2020emnlp}
Xiaoxiao Guo, Mo~Yu, Yupeng Gao, Chuang Gan, Murray Campbell, and Shiyu Chang.
  2020.
\newblock \href {https://doi.org/10.18653/v1/2020.emnlp-main.624} {Interactive
  fiction game playing as multi-paragraph reading comprehension with
  reinforcement learning}.
\newblock In \emph{Proceedings of the Conference on Empirical Methods in
  Natural Language Processing (EMNLP)}, pages 7755--7765.

\bibitem[{Hasselt et~al.(2016)Hasselt, Guez, and Silver}]{hasselt2016doubledqn}
Hado~van Hasselt, Arthur Guez, and David Silver. 2016.
\newblock Deep reinforcement learning with double q-learning.
\newblock In \emph{Proceedings of the AAAI Conference on Artificial
  Intelligence (AAAI)}, volume~30, pages 2094--2100.

\bibitem[{Hausknecht et~al.(2020)Hausknecht, Ammanabrolu, C{\^o}t{\'e}, and
  Yuan}]{hausknecht2019jericho}
Matthew Hausknecht, Prithviraj Ammanabrolu, Marc-Alexandre C{\^o}t{\'e}, and
  Xingdi Yuan. 2020.
\newblock Interactive fiction games: A colossal adventure.
\newblock In \emph{Proceedings of the AAAI Conference on Artificial
  Intelligence (AAAI)}, volume~34, pages 7903--7910.

\bibitem[{Hausknecht et~al.(2019)Hausknecht, Loynd, Yang, Swaminathan, and
  Williams}]{hausknecht2019nail}
Matthew Hausknecht, Ricky Loynd, Greg Yang, Adith Swaminathan, and Jason~D
  Williams. 2019.
\newblock Nail: A general interactive fiction agent.
\newblock \emph{arXiv preprint arXiv:1902.04259}.

\bibitem[{He et~al.(2016)He, Chen, He, Gao, Li, Deng, and
  Ostendorf}]{he2016drrn}
Ji~He, Jianshu Chen, Xiaodong He, Jianfeng Gao, Lihong Li, Li~Deng, and Mari
  Ostendorf. 2016.
\newblock \href {https://doi.org/10.18653/v1/P16-1153} {Deep reinforcement
  learning with a natural language action space}.
\newblock In \emph{Proceedings of the Annual Meeting of the Association for
  Computational Linguistics (ACL)}, pages 1621--1630.

\bibitem[{Jain et~al.(2020)Jain, Fedus, Larochelle, Precup, and
  Bellemare}]{jain2019aaai}
Vishal Jain, William Fedus, Hugo Larochelle, Doina Precup, and Marc~G
  Bellemare. 2020.
\newblock Algorithmic improvements for deep reinforcement learning applied to
  interactive fiction.
\newblock In \emph{Proceedings of the AAAI Conference on Artificial
  Intelligence (AAAI)}, volume~34, pages 4328--4336.

\bibitem[{Jiang et~al.(2019)Jiang, Gu, Murphy, and Finn}]{jiang2019hrl}
Yiding Jiang, Shixiang~Shane Gu, Kevin~P Murphy, and Chelsea Finn. 2019.
\newblock Language as an abstraction for hierarchical deep reinforcement
  learning.
\newblock In \emph{Advances in Neural Information Processing Systems
  (NeurIPS)}, volume~32, pages 9419--9431.

\bibitem[{Kaelbling(1993)}]{kaelbling1993goal}
Leslie~Pack Kaelbling. 1993.
\newblock Learning to achieve goals.
\newblock In \emph{Proceedings of the International Joint Conference on
  Artificial Intelligence (IJCAI)}, pages 1094--1098.

\bibitem[{Kulkarni et~al.(2016)Kulkarni, Narasimhan, Saeedi, and
  Tenenbaum}]{kulkarni2016hrl}
Tejas~D Kulkarni, Karthik Narasimhan, Ardavan Saeedi, and Josh Tenenbaum. 2016.
\newblock Hierarchical deep reinforcement learning: Integrating temporal
  abstraction and intrinsic motivation.
\newblock \emph{Advances in Neural Information Processing Systems (NeurIPS)},
  29:3675--3683.

\bibitem[{Luketina et~al.(2019)Luketina, Nardelli, Farquhar, Foerster, Andreas,
  Grefenstette, Whiteson, and Rocktäschel}]{luketina2019survey}
Jelena Luketina, Nantas Nardelli, Gregory Farquhar, Jakob Foerster, Jacob
  Andreas, Edward Grefenstette, Shimon Whiteson, and Tim Rocktäschel. 2019.
\newblock \href {https://doi.org/10.24963/ijcai.2019/880} {A survey of
  reinforcement learning informed by natural language}.
\newblock In \emph{Proceedings of the International Joint Conference on
  Artificial Intelligence (IJCAI)}, pages 6309--6317.

\bibitem[{Murugesan et~al.(2021)Murugesan, Atzeni, Kapanipathi, Shukla,
  Kumaravel, Tesauro, Talamadupula, Sachan, and
  Campbell}]{murugesan2020commonsense}
Keerthiram Murugesan, Mattia Atzeni, Pavan Kapanipathi, Pushkar Shukla, Sadhana
  Kumaravel, Gerald Tesauro, Kartik Talamadupula, Mrinmaya Sachan, and Murray
  Campbell. 2021.
\newblock Text-based rl agents with commonsense knowledge: New challenges,
  environments and baselines.
\newblock In \emph{Proceedings of the AAAI Conference on Artificial
  Intelligence (AAAI)}, volume~35, pages 9018--9027.

\bibitem[{Murugesan et~al.(2020)Murugesan, Atzeni, Shukla, Sachan, Kapanipathi,
  and Talamadupula}]{murugesan2020kg}
Keerthiram Murugesan, Mattia Atzeni, Pushkar Shukla, Mrinmaya Sachan, Pavan
  Kapanipathi, and Kartik Talamadupula. 2020.
\newblock Enhancing text-based reinforcement learning agents with commonsense
  knowledge.
\newblock \emph{arXiv preprint arXiv:2005.00811}.

\bibitem[{Nachum et~al.(2018)Nachum, Gu, Lee, and Levine}]{nachum2018hrl}
Ofir Nachum, Shixiang~Shane Gu, Honglak Lee, and Sergey Levine. 2018.
\newblock Data-efficient hierarchical reinforcement learning.
\newblock In \emph{Advances in Neural Information Processing Systems
  (NeurIPS)}, volume~31, pages 3303--3313.

\bibitem[{Narasimhan et~al.(2015)Narasimhan, Kulkarni, and
  Barzilay}]{narasimhan2015language}
Karthik Narasimhan, Tejas~D Kulkarni, and Regina Barzilay. 2015.
\newblock \href {https://doi.org/10.18653/v1/D15-1001} {Language understanding
  for text-based games using deep reinforcement learning}.
\newblock In \emph{Proceedings of the Conference on Empirical Methods in
  Natural Language Processing (EMNLP)}, pages 1--11.

\bibitem[{Oh et~al.(2017)Oh, Singh, Lee, and Kohli}]{oh2017zero}
Junhyuk Oh, Satinder Singh, Honglak Lee, and Pushmeet Kohli. 2017.
\newblock Zero-shot task generalization with multi-task deep reinforcement
  learning.
\newblock In \emph{International Conference on Machine Learning (ICML)},
  volume~70, pages 2661--2670.

\bibitem[{Peng et~al.(2017)Peng, Li, Li, Gao, Celikyilmaz, Lee, and
  Wong}]{peng2017HRLforDialogue}
Baolin Peng, Xiujun Li, Lihong Li, Jianfeng Gao, Asli Celikyilmaz, Sungjin Lee,
  and Kam-Fai Wong. 2017.
\newblock \href {https://doi.org/10.18653/v1/D17-1237} {Composite
  task-completion dialogue policy learning via hierarchical deep reinforcement
  learning}.
\newblock In \emph{Proceedings of the Conference on Empirical Methods in
  Natural Language Processing (EMNLP)}, pages 2231--2240.

\bibitem[{Saleh et~al.(2020)Saleh, Jaques, Ghandeharioun, Shen, and
  Picard}]{saleh2020HRLforDialogue}
Abdelrhman Saleh, Natasha Jaques, Asma Ghandeharioun, Judy Shen, and Rosalind
  Picard. 2020.
\newblock Hierarchical reinforcement learning for open-domain dialog.
\newblock In \emph{Proceedings of the AAAI Conference on Artificial
  Intelligence (AAAI)}, volume~34, pages 8741--8748.

\bibitem[{Schaul et~al.(2015)Schaul, Quan, Antonoglou, and
  Silver}]{schaul2015per}
Tom Schaul, John Quan, Ioannis Antonoglou, and David Silver. 2015.
\newblock Prioritized experience replay.
\newblock \emph{arXiv preprint arXiv:1511.05952}.

\bibitem[{Schlichtkrull et~al.(2018)Schlichtkrull, Kipf, Bloem, Van Den~Berg,
  Titov, and Welling}]{schlichtkrull2018rgcn}
Michael Schlichtkrull, Thomas~N Kipf, Peter Bloem, Rianne Van Den~Berg, Ivan
  Titov, and Max Welling. 2018.
\newblock Modeling relational data with graph convolutional networks.
\newblock In \emph{European Semantic Web Conference (ESWC)}, pages 593--607.

\bibitem[{Schulman et~al.(2017)Schulman, Wolski, Dhariwal, Radford, and
  Klimov}]{schulman2017ppo}
John Schulman, Filip Wolski, Prafulla Dhariwal, Alec Radford, and Oleg Klimov.
  2017.
\newblock Proximal policy optimization algorithms.
\newblock \emph{arXiv preprint arXiv:1707.06347}.

\bibitem[{Shiarlis et~al.(2018)Shiarlis, Wulfmeier, Salter, Whiteson, and
  Posner}]{shiarlis2018taco}
Kyriacos Shiarlis, Markus Wulfmeier, Sasha Salter, Shimon Whiteson, and Ingmar
  Posner. 2018.
\newblock Taco: Learning task decomposition via temporal alignment for control.
\newblock In \emph{International Conference on Machine Learning (ICML)},
  volume~80, pages 4654--4663.

\bibitem[{Shu et~al.(2018)Shu, Xiong, and Socher}]{shu2018hrl}
Tianmin Shu, Caiming Xiong, and Richard Socher. 2018.
\newblock \href {https://openreview.net/forum?id=SJJQVZW0b} {Hierarchical and
  interpretable skill acquisition in multi-task reinforcement learning}.
\newblock In \emph{International Conference on Learning Representations
  (ICLR)}.

\bibitem[{Silver et~al.(2016)Silver, Huang, Maddison, Guez, Sifre, Van
  Den~Driessche, Schrittwieser, Antonoglou, Panneershelvam, Lanctot
  et~al.}]{silver2016mastering}
David Silver, Aja Huang, Chris~J Maddison, Arthur Guez, Laurent Sifre, George
  Van Den~Driessche, Julian Schrittwieser, Ioannis Antonoglou, Veda
  Panneershelvam, Marc Lanctot, et~al. 2016.
\newblock Mastering the game of go with deep neural networks and tree search.
\newblock \emph{nature}, 529(7587):484--489.

\bibitem[{Sohn et~al.(2018)Sohn, Oh, and Lee}]{sohn2018subtask}
Sungryull Sohn, Junhyuk Oh, and Honglak Lee. 2018.
\newblock Hierarchical reinforcement learning for zero-shot generalization with
  subtask dependencies.
\newblock In \emph{Advances in Neural Information Processing Systems
  (NeurIPS)}, volume~31, pages 7156--7166.

\bibitem[{Sutton et~al.(1999)Sutton, Precup, and Singh}]{sutton1999hrl}
Richard~S Sutton, Doina Precup, and Satinder Singh. 1999.
\newblock Between mdps and semi-mdps: A framework for temporal abstraction in
  reinforcement learning.
\newblock \emph{Artificial intelligence}, 112(1-2):181--211.

\bibitem[{Urbanek et~al.(2019)Urbanek, Fan, Karamcheti, Jain, Humeau, Dinan,
  Rockt{\"a}schel, Kiela, Szlam, and Weston}]{urbanek2019light}
Jack Urbanek, Angela Fan, Siddharth Karamcheti, Saachi Jain, Samuel Humeau,
  Emily Dinan, Tim Rockt{\"a}schel, Douwe Kiela, Arthur Szlam, and Jason
  Weston. 2019.
\newblock \href {https://doi.org/10.18653/v1/D19-1062} {Learning to speak and
  act in a fantasy text adventure game}.
\newblock In \emph{Proceedings of the Conference on Empirical Methods in
  Natural Language Processing (EMNLP)}, pages 673--683.

\bibitem[{Vaswani et~al.(2017)Vaswani, Shazeer, Parmar, Uszkoreit, Jones,
  Gomez, Kaiser, and Polosukhin}]{vaswani2017transformer}
Ashish Vaswani, Noam Shazeer, Niki Parmar, Jakob Uszkoreit, Llion Jones,
  Aidan~N Gomez, {\L}ukasz Kaiser, and Illia Polosukhin. 2017.
\newblock Attention is all you need.
\newblock In \emph{Advances in Neural Information Processing Systems
  (NeurIPS)}, volume~30, pages 5998--6008.

\bibitem[{Vezhnevets et~al.(2017)Vezhnevets, Osindero, Schaul, Heess,
  Jaderberg, Silver, and Kavukcuoglu}]{vezhnevets2017hrl_feudal}
Alexander~Sasha Vezhnevets, Simon Osindero, Tom Schaul, Nicolas Heess, Max
  Jaderberg, David Silver, and Koray Kavukcuoglu. 2017.
\newblock {F}e{U}dal networks for hierarchical reinforcement learning.
\newblock In \emph{International Conference on Machine Learning (ICML)},
  volume~70, pages 3540--3549.

\bibitem[{Vinyals et~al.(2019)Vinyals, Babuschkin, Czarnecki, Mathieu, Dudzik,
  Chung, Choi, Powell, Ewalds, Georgiev et~al.}]{vinyals2019alphastar}
Oriol Vinyals, Igor Babuschkin, Wojciech~M Czarnecki, Micha{\"e}l Mathieu,
  Andrew Dudzik, Junyoung Chung, David~H Choi, Richard Powell, Timo Ewalds,
  Petko Georgiev, et~al. 2019.
\newblock Grandmaster level in starcraft ii using multi-agent reinforcement
  learning.
\newblock \emph{Nature}, 575(7782):350--354.

\bibitem[{Xu et~al.(2020{\natexlab{a}})Xu, Chen, Fang, Wang, and
  Zhang}]{xu2020cog}
Yunqiu Xu, Ling Chen, Meng Fang, Yang Wang, and Chengqi Zhang.
  2020{\natexlab{a}}.
\newblock Deep reinforcement learning with transformers for text adventure
  games.
\newblock In \emph{IEEE Conference on Games (CoG)}, pages 65--72.

\bibitem[{Xu et~al.(2020{\natexlab{b}})Xu, Fang, Chen, Du, Zhou, and
  Zhang}]{xu2020nips}
Yunqiu Xu, Meng Fang, Ling Chen, Yali Du, Joey~Tianyi Zhou, and Chengqi Zhang.
  2020{\natexlab{b}}.
\newblock Deep reinforcement learning with stacked hierarchical attention for
  text-based games.
\newblock In \emph{Advances in Neural Information Processing Systems
  (NeurIPS)}, volume~33, pages 16495--16507.

\bibitem[{Yao et~al.(2020)Yao, Rao, Hausknecht, and Narasimhan}]{yao2020emnlp}
Shunyu Yao, Rohan Rao, Matthew Hausknecht, and Karthik Narasimhan. 2020.
\newblock \href {https://doi.org/10.18653/v1/2020.emnlp-main.704} {Keep {CALM}
  and explore: Language models for action generation in text-based games}.
\newblock In \emph{Proceedings of the Conference on Empirical Methods in
  Natural Language Processing (EMNLP)}, pages 8736--8754.

\bibitem[{Yin and May(2019)}]{yin2019cog}
Xusen Yin and Jonathan May. 2019.
\newblock Comprehensible context-driven text game playing.
\newblock \emph{IEEE Conference on Games (CoG)}, pages 1--8.

\bibitem[{Yin et~al.(2020)Yin, Weischedel, and May}]{yin2020emnlpfinding}
Xusen Yin, Ralph Weischedel, and Jonathan May. 2020.
\newblock \href {https://doi.org/10.18653/v1/2020.findings-emnlp.273} {Learning
  to generalize for sequential decision making}.
\newblock In \emph{Proceedings of the Conference on Empirical Methods in
  Natural Language Processing: Findings (EMNLP-Findings)}, pages 3046--3063.

\bibitem[{Yuan et~al.(2018)Yuan, Côté, Sordoni, Laroche, Tachet~des Combes,
  Hausknecht, and Trischler}]{yuan2018counting}
Xingdi~(Eric) Yuan, Marc-Alexandre Côté, Alessandro Sordoni, Romain Laroche,
  Remi Tachet~des Combes, Matthew Hausknecht, and Adam Trischler. 2018.
\newblock Counting to explore and generalize in text-based games.
\newblock In \emph{European Workshop on Reinforcement Learning (EWRL)}.

\bibitem[{Zahavy et~al.(2018)Zahavy, Haroush, Merlis, Mankowitz, and
  Mannor}]{zahavy2018nips}
Tom Zahavy, Matan Haroush, Nadav Merlis, Daniel~J Mankowitz, and Shie Mannor.
  2018.
\newblock Learn what not to learn: Action elimination with deep reinforcement
  learning.
\newblock In \emph{Advances in Neural Information Processing Systems
  (NeurIPS)}, volume~31, pages 3562--3573.

\bibitem[{Zhang et~al.(2020)Zhang, Xu, Wang, Wu, Keutzer, Gonzalez, and
  Tian}]{zhang2020counting}
Tianjun Zhang, Huazhe Xu, Xiaolong Wang, Yi~Wu, Kurt Keutzer, Joseph~E
  Gonzalez, and Yuandong Tian. 2020.
\newblock Bebold: Exploration beyond the boundary of explored regions.
\newblock \emph{arXiv preprint arXiv:2012.08621}.

\end{thebibliography}

\clearpage
\onecolumn
\appendix
The appendix is organized as follows: Sec. \ref{appendix_pomdp}
formalizes the text-based games as POMDP. Sec. \ref{appendix_goal_set_generator} details the goal set generator. Sec. \ref{appendix_more_exp_results} provides more experimental results. Sec. \ref{appendix_more_game_examples} provides more examples of the observations and the interaction process. 

\section{Text-based games as POMDP \label{appendix_pomdp}}
Text-based games can be formulated as Partially Observable Markov Decision Processes (POMDPs), which can be defined as a 7-tuple: $\langle \mathcal{S}, \mathcal{A}, T, R, \Omega, O, \gamma \rangle$, where $\mathcal{S}$ denotes the state set, $\mathcal{A}$ denotes the action set, $T$ denotes the state transition probabilities, $R$ denotes the reward function, $\Omega$ denotes the observation set, $O$ denotes the conditional observation probabilities and $\gamma \in (0,1]$ denotes the discount factor. 
At each time step, the agent will receive an observation $o_t \in \Omega$, depending on the current state and previous action via the conditional observation probability $O(o_t|s_t,a_{t-1})$. By executing an action $a_t \in \mathcal{A}$, the environment will transit into a new state based on the state transition probability $T(s_{t+1}|s_t,a_t)$, and the agent will receive the reward $r_{t+1} = R(s_t,a_t)$. Same as MDPs, the goal of the agent is to learn an optimal policy $\pi^*$ to maximize the expected future discounted sum of rewards from each time step:  $R_t = \mathbb{E}[\sum_{k=0}^{\infty}\gamma^k r_{t+k+1}]$.
Instead of using raw textual-based observations, in this work we consider the KG-based observation $o_t^{\text{KG}}$. 
Besides, we select the action $a_t$ from the admissible action set $A_t$, which is a subset of $\mathcal{A}$. Both $o_t^{\text{KG}}$ and $A_t$ are provided by the environment for each time step.

\section{Goal set generator \label{appendix_goal_set_generator}}
We design a simple non-learning-based goal set generator by exploiting the KG-based observation in the cooking theme.  
Algorithm \ref{algo_GoalSetGeneration} shows the pipeline for obtaining the goal set $\mathcal{G}_t$. We first obtain the ingredient set $\mathcal{I}$. For each ingredient $i \in \mathcal{I}$, we first check whether it has been collected, then obtain its status set $\mathcal{S}_i$ and requirement set $\mathcal{R}_i$. 
We consider three types of goals: 1) ``find'' requires the agent to find and collect an uncollected ingredient, 2) ``prepare'' requires the agent to prepare an ingredient to satisfy a requirement, and 3) ``eat'', that the agent is required to prepare and eat the final meal. 
Algorithm \ref{algo_GoalReward} shows the pipeline for assigning the goal-conditioned reward $r_t^{\text{goal}}$. We first obtain the type of a goal $g$, then check whether this goal has been accomplished given $a_t$ and $o_{t+1}^{\text{KG}}$.  Some functions in Algorithm \ref{algo_GoalSetGeneration} can be reused here.  $r_t^{\text{goal}}$ is a binary reward that we will assign $r_t^{\text{goal}} = r_{\text{max}}$ if $g$ is accomplished successfully, otherwise $r_{\text{min}}$ (still not finished, or failed). Algorithm \ref{algo_GoalSetGeneration} and Algorithm \ref{algo_GoalReward} can also be implemented via learning-based methods. For example, the functions can be achieved by a QA model by answering specific questions. 

\clearpage
\begin{algorithm}[t!]
\small
\caption{Goal set generation}
\label{algo_GoalSetGeneration}
\textbf{Input:} Knowledge graph $o_t^{\text{KG}}$ \\
\textbf{Initialize:} Goal set $\mathcal{G}_t \leftarrow \emptyset$
\begin{algorithmic}[1]
\State Get ingredient set $\mathcal{I} \leftarrow \text{GetIng}(o_t^{\text{KG}})$
\For{each ingredient $i \in \mathcal{I}$}
\If {$\text{CheckCollection}(i, o_t^{\text{KG}}) = $ False}
\State Add goal ``find $i$'' to $\mathcal{G}_t$
\Else
    \State Get status set $\mathcal{S}_i \leftarrow \text{GetStatus}(i,o_t^{\text{KG}})$
    \State Get requirement set $\mathcal{R}_i \leftarrow \text{GetReq}(i,o_t^{\text{KG}})$
    \For{each requirement $r_i \in \mathcal{R}_i$}
        \If {$r_i \notin \mathcal{S}_i$}
        \State Add goal ``$r_i$ $i$'' to $\mathcal{G}_t$
        \EndIf
    \EndFor
\EndIf
\EndFor
\If {$\mathcal{G}_t = \emptyset$}
\State Add goal ``prepare and eat meal'' to $\mathcal{G}_t$
\EndIf
\State \textbf{return} $\mathcal{G}_t$.
\end{algorithmic}
\end{algorithm}

\begin{algorithm}[t!]
\small
\caption{Goal-conditioned reward acquisition}
\label{algo_GoalReward}
\textbf{Input:} Knowledge graph $o_{t+1}^{\text{KG}}$, goal $g$, rewards $\{ r_{\text{min}}, r_{\text{max}} \}$ \\
\textbf{Initialize:} $\text{FLAG} \leftarrow$ False
\begin{algorithmic}[1]
\State Obtain goal type $g_{\text{type}} \leftarrow \text{GetGoalType}(g)$
\If{$g_{\text{type}} = $ ``find''}
    \State Get ingredient $i$ from $g$
    \If{$\text{CheckCollection}(i, o_{t+1}^{\text{KG}}) =$ True}
        \State $\text{FLAG} \leftarrow$ True
    \EndIf
\ElsIf{$g_{\text{type}} = $ ``prepare''}
    \State Get ingredient $i$, requirement $r$ from $g$
    \State Get status set $\mathcal{S}_i \leftarrow \text{GetStatus}(i, o_{t+1}^{\text{KG}})$
    \If {$r \in \mathcal{S}_i$}
        \State $\text{FLAG} \leftarrow$ True
    \EndIf
\Else
    \If{$\text{CheckExistance}(\text{``meal''}, o_{t+1}^{\text{KG}}) =$ True}
        \State $\text{FLAG} \leftarrow$ True
    \EndIf
\EndIf
\If {$\text{FLAG} = $ True}
    \State \textbf{return} $r_{\text{max}}$.
\Else
    \State \textbf{return} $r_{\text{min}}$.
\EndIf
\end{algorithmic}
\end{algorithm}

\clearpage

\section{More experiment results \label{appendix_more_exp_results}}

\begin{figure*}[htb]
\centering
\includegraphics[width=0.95\textwidth]{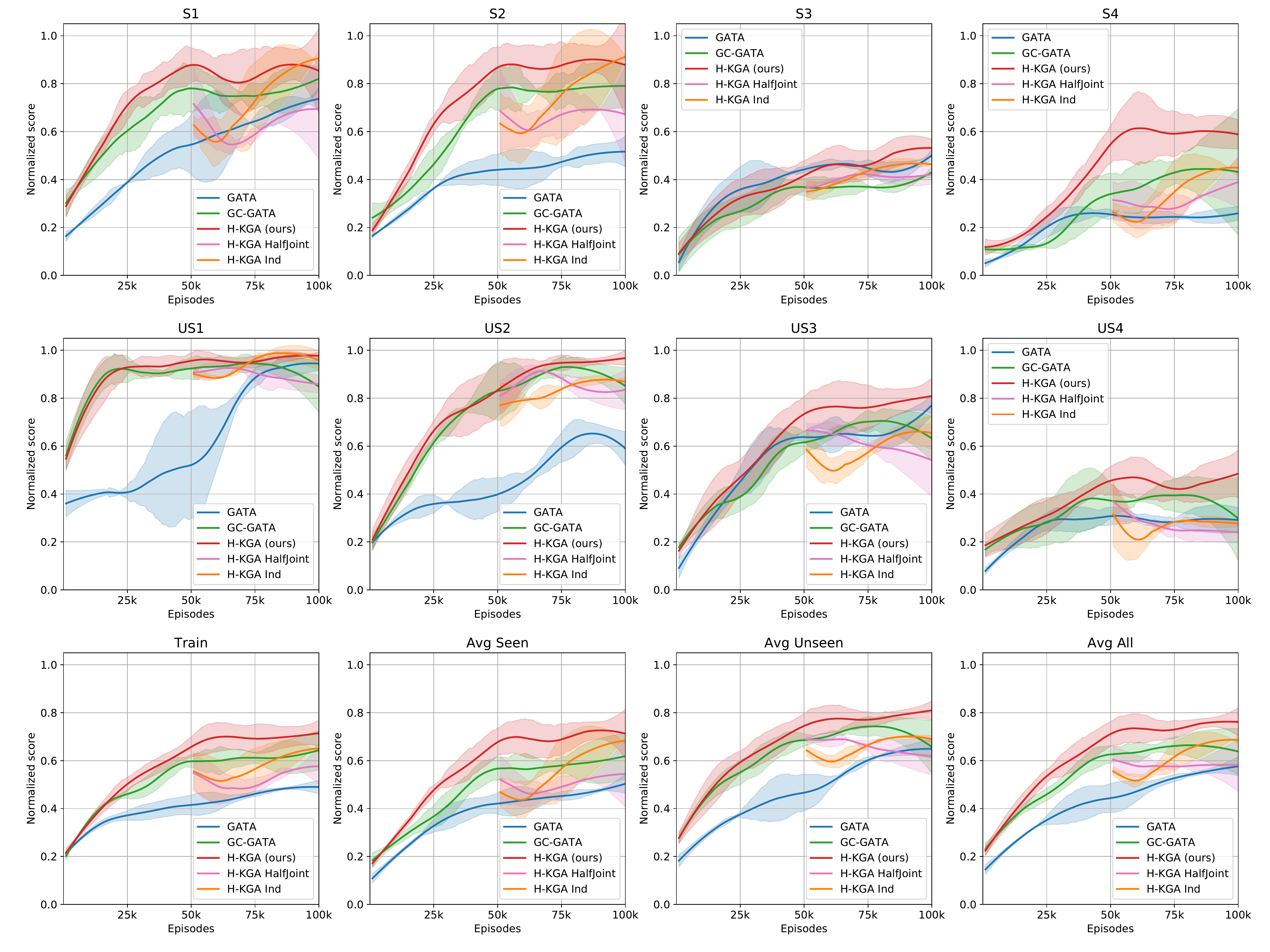}
\caption{The full results for Sec. \ref{section_exp_main_result}. }
\label{exp1sup_main}
\end{figure*}

\begin{figure*}[htb]
\centering
\includegraphics[width=0.95\textwidth]{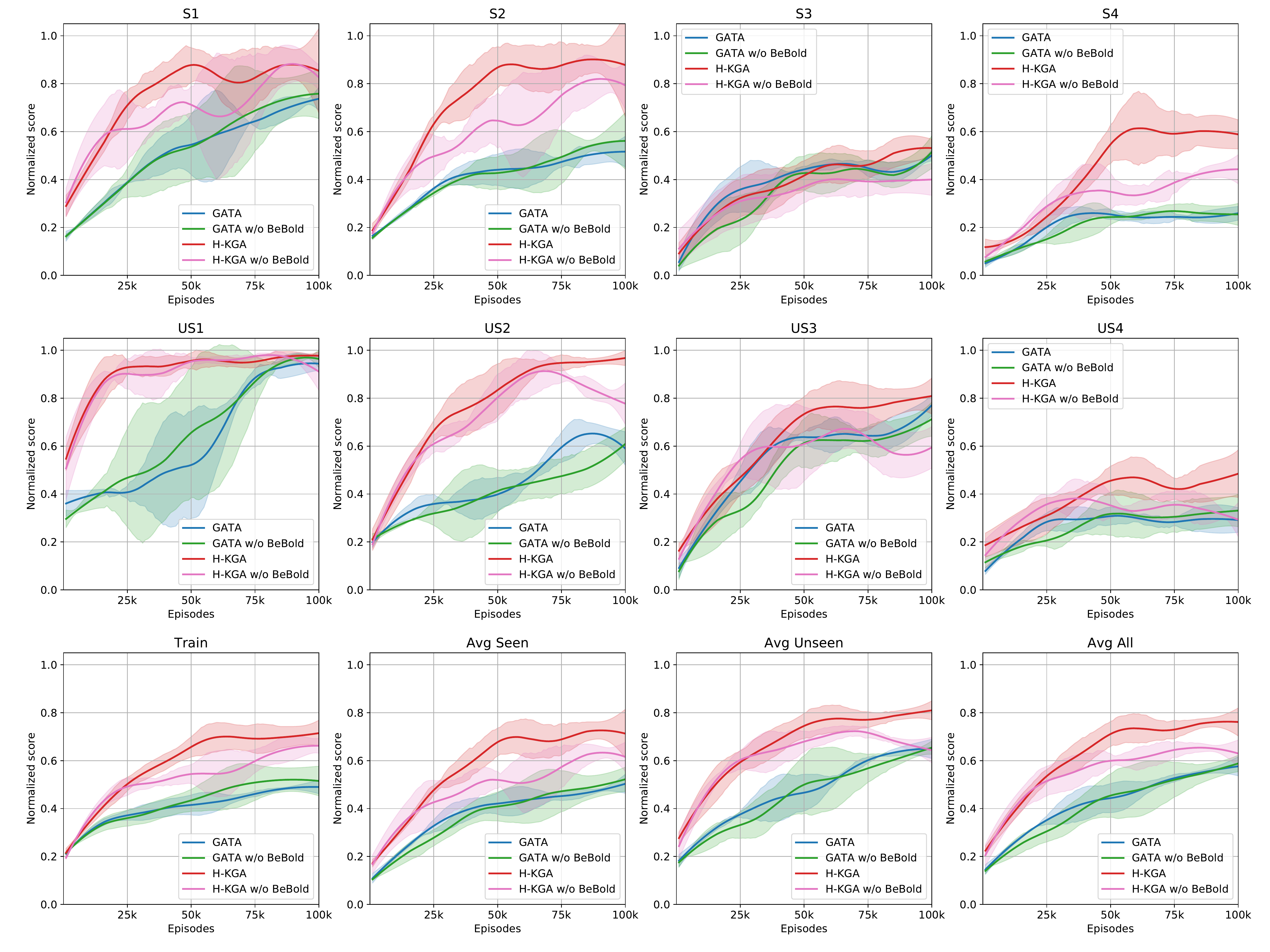}
\caption{The full results for Sec. \ref{section_exp_bebold}.}
\label{exp2sup_woBeBold}
\end{figure*}

\begin{figure*}[htb]
\centering
\includegraphics[width=0.95\textwidth]{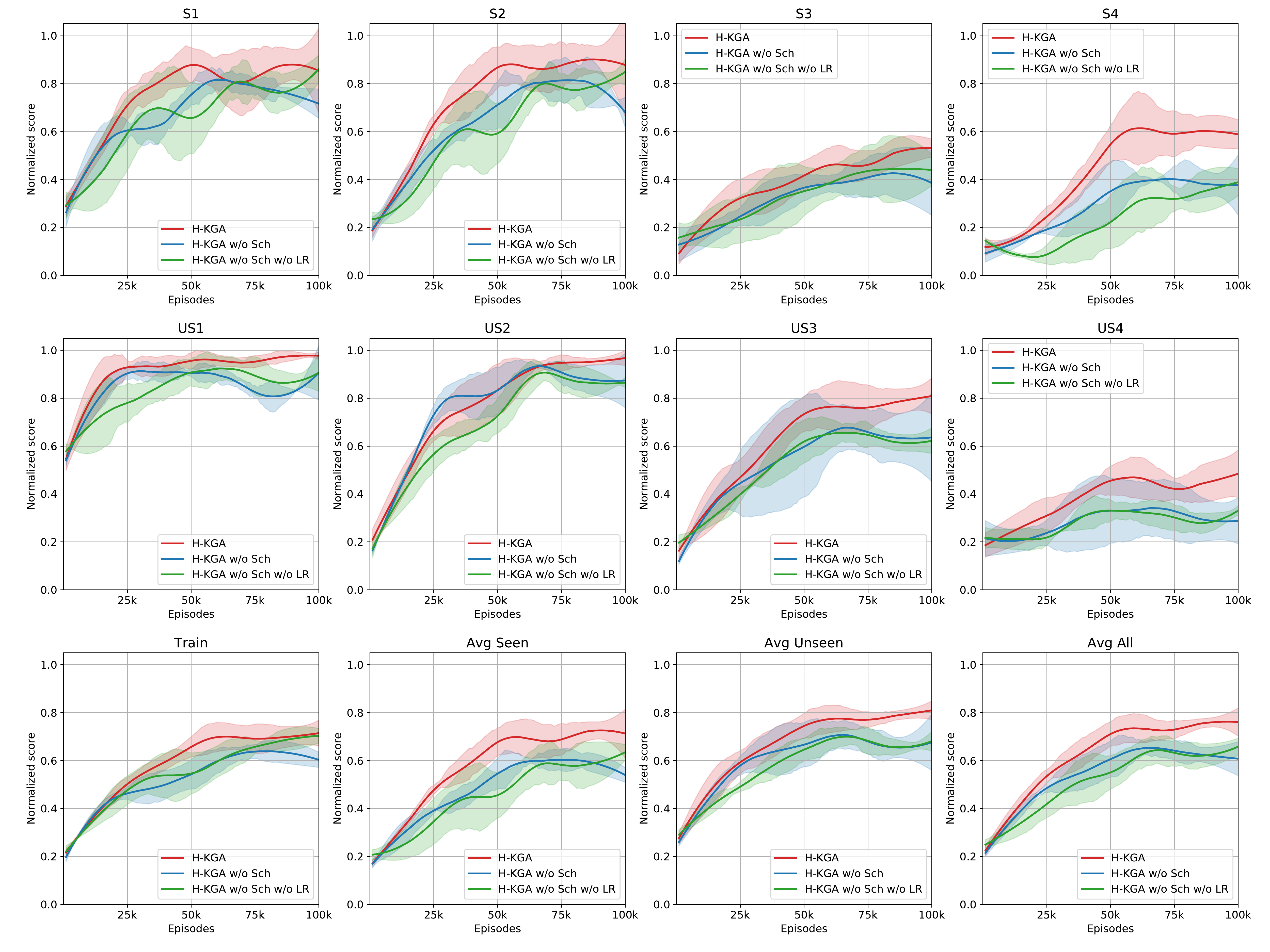}
\caption{The full results for Sec. \ref{section_exp_mtl}. }
\label{exp3sup_woMTL}
\end{figure*}

\begin{figure*}[htb]
\centering
\includegraphics[width=0.95\textwidth]{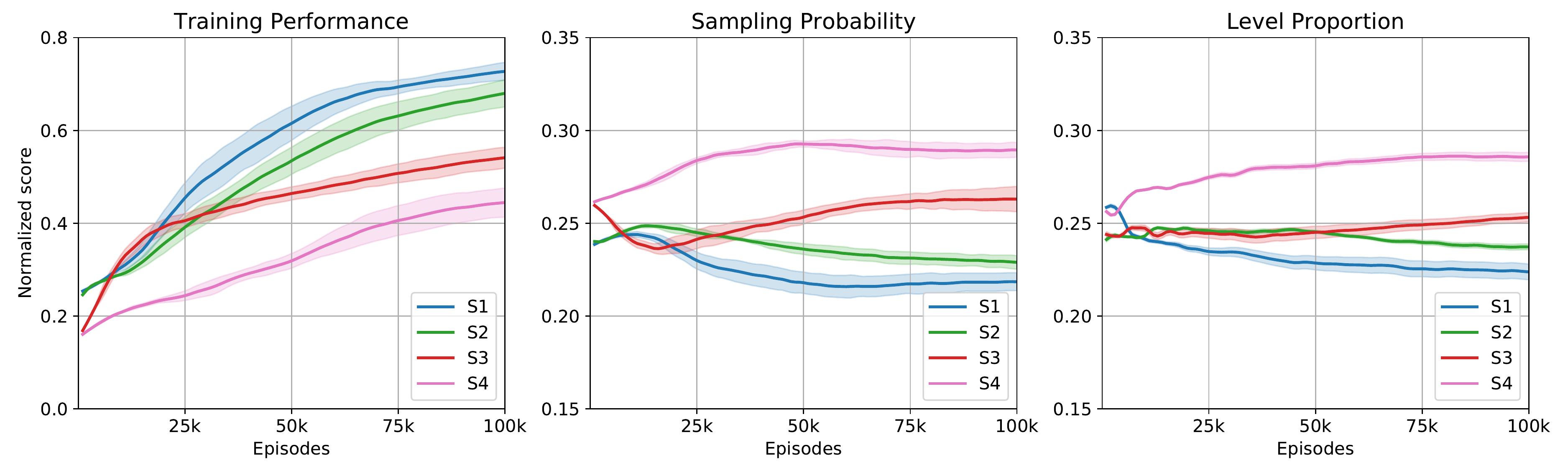}
\caption{H-KGA's training performance (left), sampling probabilities (middle) and the proportion of training levels (right). }
\label{exp4sup_sampling}
\end{figure*}

Table. \ref{table_result_sup}, Fig. \ref{exp1sup_main}, Fig. \ref{exp2sup_woBeBold} and Fig. \ref{exp3sup_woMTL} show the full results in the main paper. 
Table. \ref{table_result_sup_best} shows the maximum testing performance of models throughout the training process. 
Fig. \ref{exp4sup_sampling} shows H-KGA's training performance, the sampling probabilities and the proportion of encountered training levels.

\begin{table*}[htb]
\caption{\label{table_result_sup} 
The testing performance of models at the end of training.}
\centering
\resizebox{1.0\textwidth}{!}{
\begin{tabular}{l|cccc|cccc|ccc}
\hline 
\textbf{Model}       & \textbf{S1} & \textbf{S2} & \textbf{S3} & \textbf{S4} & \textbf{US1} & \textbf{US2}& \textbf{US3} & \textbf{US4}& \textbf{Avg Seen} & \textbf{Avg Unseen} & \textbf{Avg All} \\
\hline
GATA                 &0.70$\pm$0.08&0.43$\pm$0.02&0.48$\pm$0.04&0.28$\pm$0.05&0.90$\pm$0.04&0.59$\pm$0.08&0.71$\pm$0.11&0.28$\pm$0.09&0.47$\pm$0.04&0.62$\pm$0.07&0.55$\pm$0.06    \\
GC-GATA              &0.68$\pm$0.18&0.73$\pm$0.22&0.43$\pm$0.04&0.33$\pm$0.23&0.78$\pm$0.15&0.79$\pm$0.12&0.59$\pm$0.11&0.28$\pm$0.14&0.54$\pm$0.13&0.61$\pm$0.12&0.58$\pm$0.12    \\
\textbf{H-KGA (ours)}&\textbf{0.93}$\pm$0.06&0.85$\pm$0.17&0.50$\pm$0.05&\textbf{0.61}$\pm$0.18&0.97$\pm$0.03&\textbf{0.97}$\pm$0.03&\textbf{0.78}$\pm$0.03&\textbf{0.44}$\pm$0.12&\textbf{0.72}$\pm$0.04&\textbf{0.79}$\pm$0.04&\textbf{0.76}$\pm$0.03\\
H-KGA HalfJoint      &0.80$\pm$0.12&0.75$\pm$0.12&0.43$\pm$0.07&0.28$\pm$0.11&0.82$\pm$0.02&0.81$\pm$0.08&0.48$\pm$0.16&0.19$\pm$0.03&0.56$\pm$0.11&0.57$\pm$0.07&0.57$\pm$0.09    \\
H-KGA Ind            &0.91$\pm$0.00&\textbf{0.92}$\pm$0.04&0.49$\pm$0.02&0.47$\pm$0.04&0.95$\pm$0.07&0.88$\pm$0.00&0.64$\pm$0.04&0.27$\pm$0.03&0.70$\pm$0.02&0.68$\pm$0.01&0.69$\pm$0.02    \\
\hline
GATA w/o BeBold      &0.78$\pm$0.11&0.62$\pm$0.13&0.50$\pm$0.12&0.24$\pm$0.08&\textbf{0.99}$\pm$0.02&0.65$\pm$0.06&0.73$\pm$0.10&0.36$\pm$0.07&0.54$\pm$0.06&0.68$\pm$0.02&0.61$\pm$0.03    \\
H-KGA w/o BeBold     &0.73$\pm$0.14&0.74$\pm$0.07&0.43$\pm$0.11&0.37$\pm$0.09&0.90$\pm$0.14&0.77$\pm$0.01&0.64$\pm$0.10&0.31$\pm$0.16&0.57$\pm$0.07&0.65$\pm$0.09&0.61$\pm$0.07    \\
\hline
H-KGA w/o Sch        &0.73$\pm$0.19&0.60$\pm$0.16&0.39$\pm$0.08&0.34$\pm$0.04&0.91$\pm$0.01&0.80$\pm$0.13&0.52$\pm$0.09&0.29$\pm$0.07&0.52$\pm$0.07&0.63$\pm$0.07&0.57$\pm$0.05    \\
H-KGA w/o Sch w/o LR &0.85$\pm$0.19&0.81$\pm$0.14&\textbf{0.54}$\pm$0.10&0.33$\pm$0.15&0.82$\pm$0.23&0.79$\pm$0.17&0.60$\pm$0.16&0.30$\pm$0.10&0.63$\pm$0.14&0.63$\pm$0.16&0.63$\pm$0.15    \\
\hline
\end{tabular}
}
\end{table*}

\begin{table*}[htb]
\caption{\label{table_result_sup_best} 
The maximum testing performance of models throughout the training process. }
\centering
\resizebox{1.0\textwidth}{!}{
\begin{tabular}{l|cccc|cccc|ccc}
\hline 
\textbf{Model}       & \textbf{S1} & \textbf{S2} & \textbf{S3} & \textbf{S4} & \textbf{US1} & \textbf{US2}& \textbf{US3} & \textbf{US4}& \textbf{Avg Seen} & \textbf{Avg Unseen} & \textbf{Avg All} \\
\hline
GATA                 &0.83$\pm$0.02 &0.62$\pm$0.06 &0.58$\pm$0.02 &0.34$\pm$0.02 &0.97$\pm$0.03 &0.74$\pm$0.08 &0.82$\pm$0.02 &0.38$\pm$0.03 & 0.53$\pm$0.03     & 0.67$\pm$0.02       & 0.59$\pm$0.02    \\
GC-GATA              &0.99$\pm$0.01 &0.99$\pm$0.02 &0.52$\pm$0.02 &0.73$\pm$0.13 &\textbf{1.00}$\pm$0.00 &\textbf{1.00}$\pm$0.00 &0.82$\pm$0.03 &0.62$\pm$0.07 & 0.76$\pm$0.05     & 0.85$\pm$0.02       & 0.80$\pm$0.03    \\
\textbf{H-KGA (ours)}&\textbf{1.00}$\pm$0.00 &\textbf{1.00}$\pm$0.00 &\textbf{0.63}$\pm$0.06 &\textbf{0.89}$\pm$0.04 &\textbf{1.00}$\pm$0.00 &\textbf{1.00}$\pm$0.00 &0.89$\pm$0.04 &\textbf{0.65}$\pm$0.09 &\textbf{0.86}$\pm$0.03&\textbf{0.88}$\pm$0.02& \textbf{0.85}$\pm$0.02\\
H-KGA HalfJoint      &0.97$\pm$0.04 &0.99$\pm$0.02 &0.52$\pm$0.01 &0.69$\pm$0.11 &\textbf{1.00}$\pm$0.00 &\textbf{1.00}$\pm$0.00 &0.85$\pm$0.03 &0.61$\pm$0.08 & 0.75$\pm$0.04     & 0.84$\pm$0.04       & 0.78$\pm$0.04    \\
H-KGA Ind            &0.97$\pm$0.04 &0.99$\pm$0.02 &0.49$\pm$0.01 &0.68$\pm$0.11 &\textbf{1.00}$\pm$0.00 &\textbf{1.00}$\pm$0.00 &0.82$\pm$0.06 &0.58$\pm$0.12 & 0.72$\pm$0.04     & 0.81$\pm$0.06       & 0.75$\pm$0.04    \\
\hline
GATA w/o BeBold      &0.88$\pm$0.11 &0.71$\pm$0.08 &0.58$\pm$0.05 &0.36$\pm$0.04 &\textbf{1.00}$\pm$0.00 &0.72$\pm$0.06 &0.85$\pm$0.08 &0.50$\pm$0.07 & 0.58$\pm$0.05     & 0.73$\pm$0.04       & 0.63$\pm$0.04    \\
H-KGA w/o BeBold     &\textbf{1.00}$\pm$0.00 &0.99$\pm$0.01 &0.55$\pm$0.05 &0.72$\pm$0.04 &\textbf{1.00}$\pm$0.00 &\textbf{1.00}$\pm$0.00 &0.82$\pm$0.04 &0.61$\pm$0.05 & 0.78$\pm$0.02     & 0.80$\pm$0.02       & 0.77$\pm$0.02    \\
\hline
H-KGA w/o Sch        &\textbf{1.00}$\pm$0.00 &\textbf{1.00}$\pm$0.00 &0.57$\pm$0.06 &0.73$\pm$0.11 &\textbf{1.00}$\pm$0.00 &\textbf{1.00}$\pm$0.00 &\textbf{0.90}$\pm$0.08&0.56$\pm$0.15 & 0.77$\pm$0.04     & 0.84$\pm$0.04       & 0.79$\pm$0.03    \\
H-KGA w/o Sch w/o LR &\textbf{1.00}$\pm$0.00 &\textbf{1.00}$\pm$0.00 &\textbf{0.63}$\pm$0.12 &0.67$\pm$0.10 &\textbf{1.00}$\pm$0.00 &\textbf{1.00}$\pm$0.00 &0.82$\pm$0.05 &0.50$\pm$0.05 & 0.77$\pm$0.05     & 0.80$\pm$0.02       & 0.77$\pm$0.03    \\
\hline
\end{tabular}
}
\end{table*}

\clearpage
\section{More game examples \label{appendix_more_game_examples}}

\subsection{Observation visualization \label{app_app_observation_visualization}}

\begin{figure*}[t!]
\centering
\includegraphics[width=0.95\textwidth]{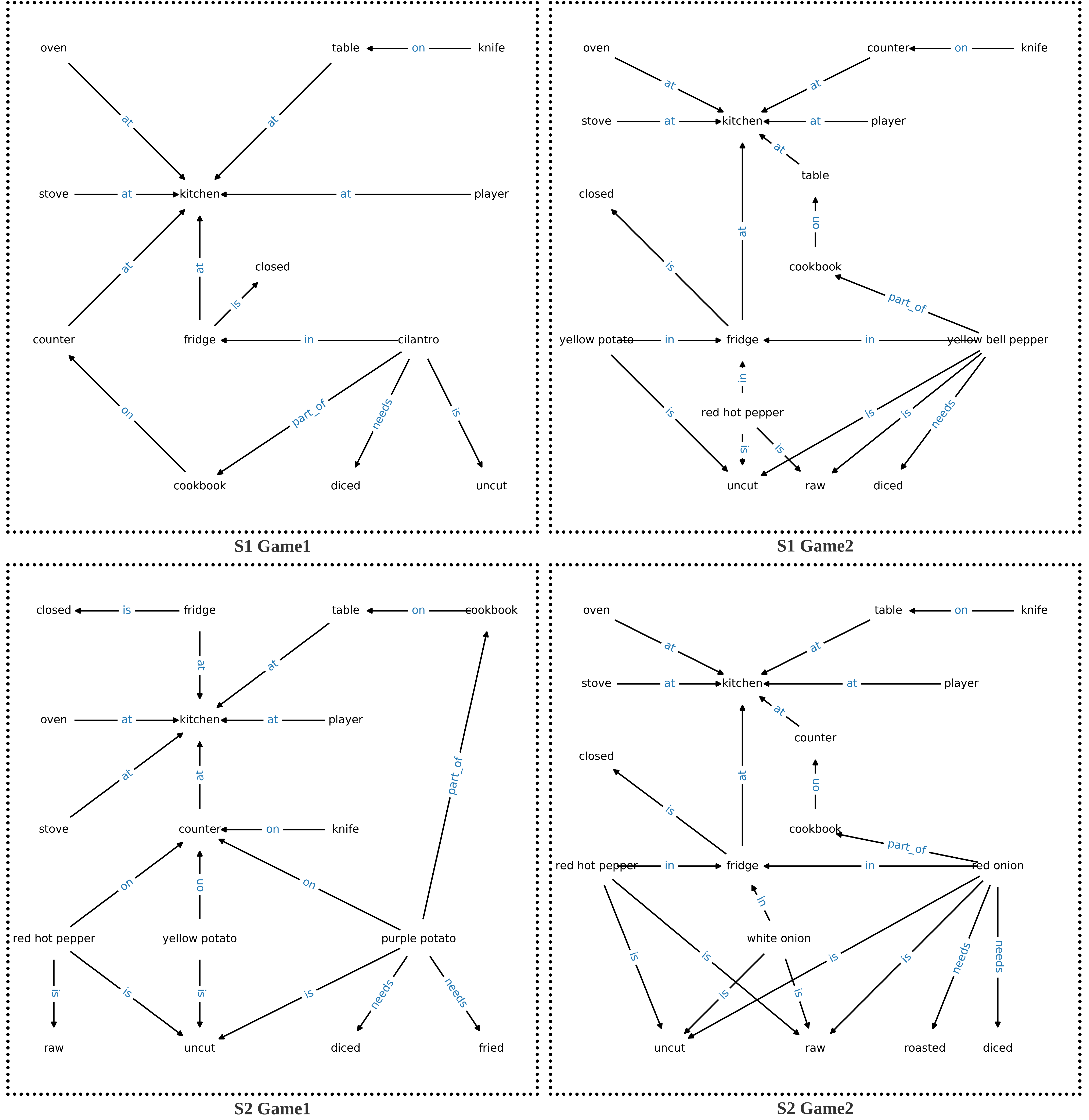}
\caption{The initial observation of four games, where ``S1 Game1'' and ``S1 Game2'' belong to level ``S1'', ``S2 Game1'' and ``S2 Game2'' belong to level ``S2''.}
\label{obs_example_1-2}
\end{figure*}

\begin{figure*}[t!]
\centering
\includegraphics[width=0.95\textwidth]{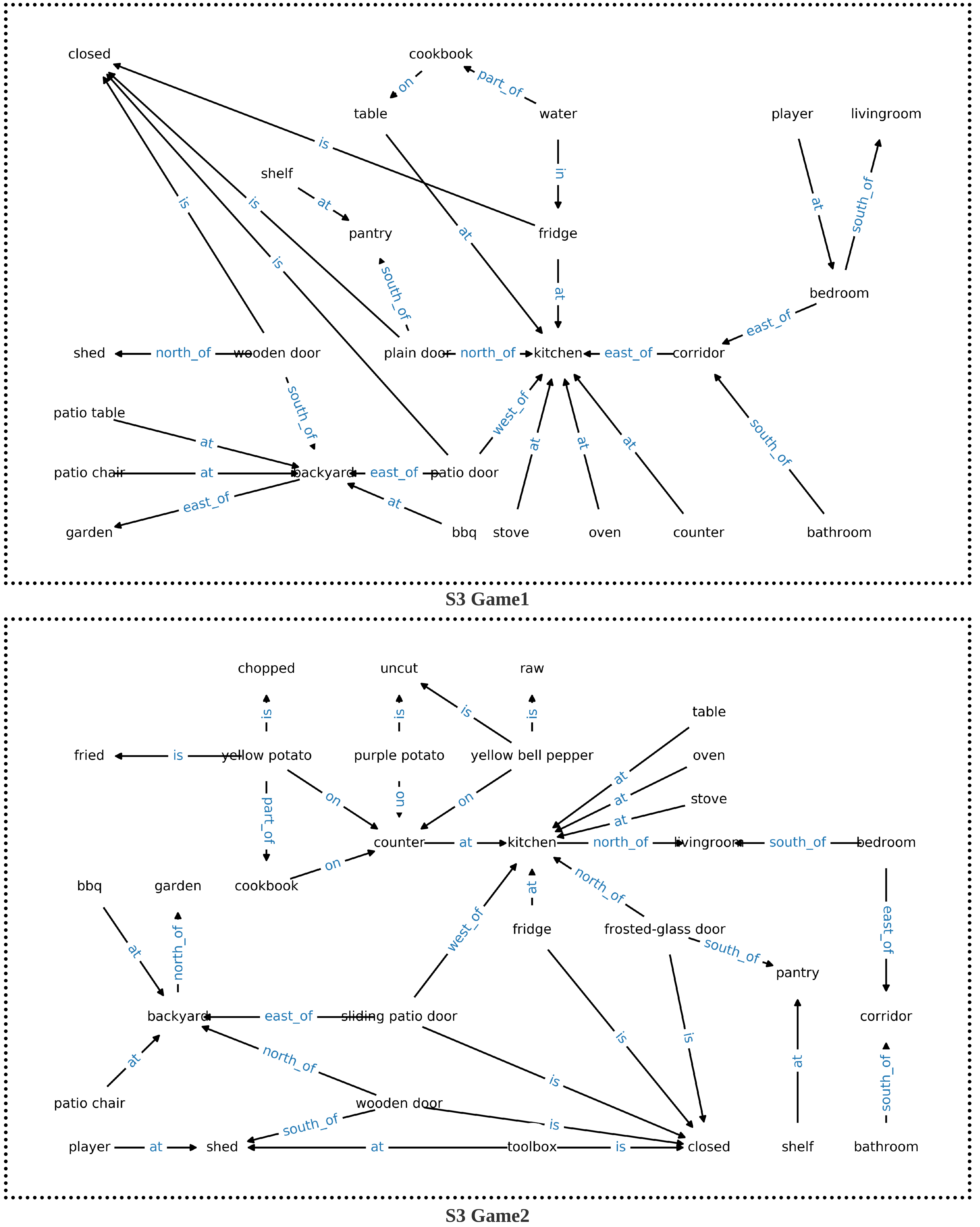}
\caption{The initial observation of two games belonging to level ``S3''. }
\label{obs_example_3}
\end{figure*}

\begin{figure*}[t!]
\centering
\includegraphics[width=0.95\textwidth]{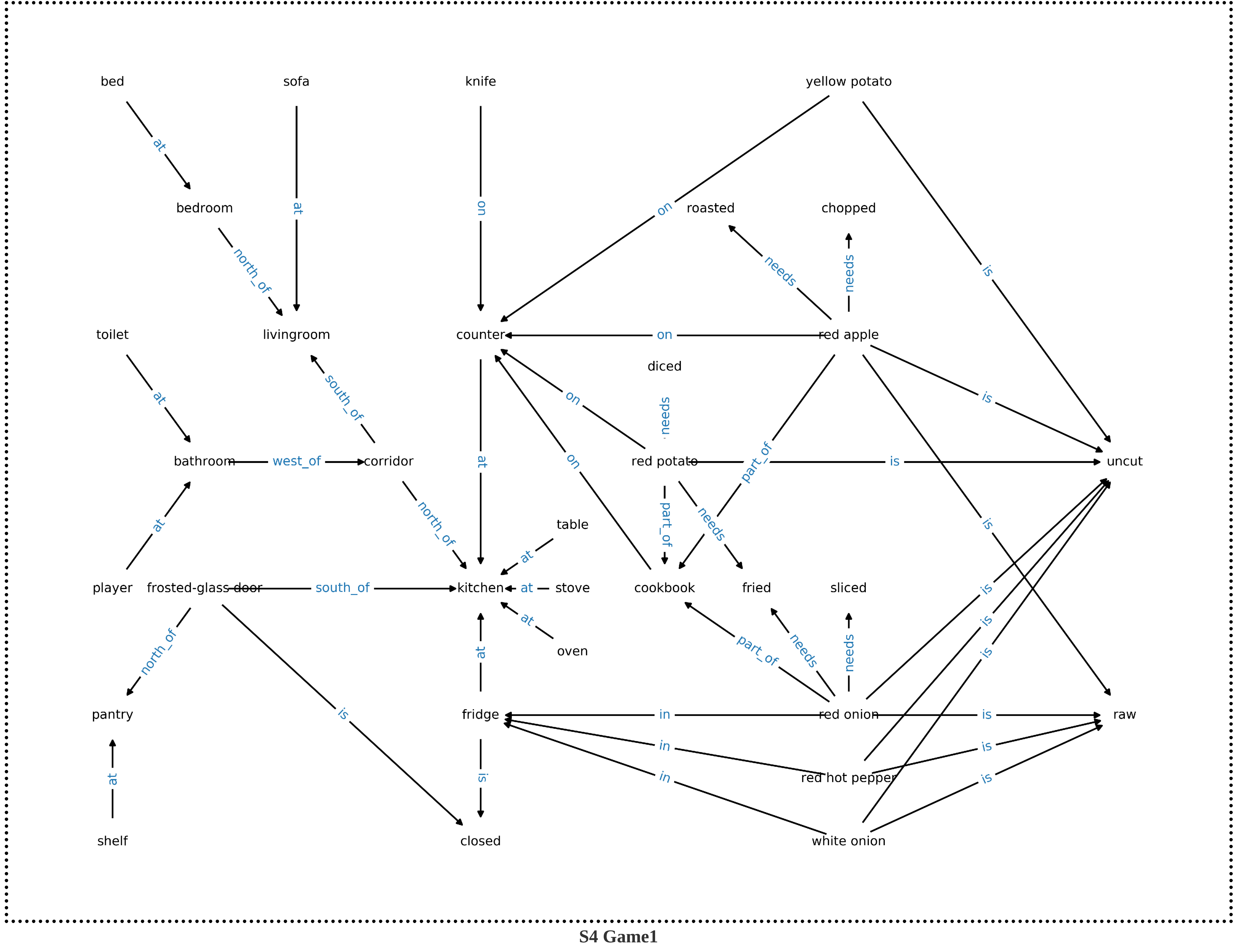}
\caption{The initial observation of one game belonging to level ``S4''. }
\label{obs_example_4}
\end{figure*}

We provide more game examples to demonstrate the difference in layouts and difficulty levels. 
Fig. \ref{obs_example_1-2} visualizes the initial observation $o_0^{\text{KG}}$ for four games, where ``S1 Game1'' and ``S1 Game2'' belong to level ``S1'', ``S2 Game1'' and ``S2 Game2'' belong to level ``S2''. 
Fig. \ref{obs_example_3} visualizes the initial observation of two games belonging to level ``S3''. Fig. \ref{obs_example_4} visualizes the initial observation of one game belonging to level ``S4''.
Games within the same level have the same complexity, but are different in their layouts. For example, ``S2 Game1'' and ``S2 Game2'' have the same number of rooms, number of ingredients and number of requirements, but the 
ingredient and requirement are different. 
Similarly, ``S3 Game1'' and ``S3 Game2'' have the same number of rooms, but are with different room connectivity. 
Games within different levels have different complexities, and their layouts are naturally different (e.g., ``S1 Game1'' v.s., ``S2 Game1'' v.s., ``S3 Game1'' v.s., ``S4 Game1''). 
The TextWorld also provides other game themes, such as the default House theme, the Coin Collector theme and the Treasure Hunter theme. 
In these themes, the agent needs to go through the rooms to find either the coin, or an object specified at the beginning of an episode. 
We believe that the Cooking theme we use in this work is able to cover these themes, as it requires the agent to navigate through different numbers of rooms, collect different numbers of ingredients, and prepare them in different ways.

\clearpage
\subsection{Interaction \label{app_app_interaction}}

In this section, we show the walk-through interactions of two games: ``S1 Game1'', which belongs to level ``S1'', and ``S4 Game1'', which belongs to level ``S4''. 
In ``S1 Game1'', the player as well as all the objects are in one room  (``kitchen''), and there's only one ingredient (``cilantro'') and one requirement (``diced'').
In ``S4 Game1'', there are multiple rooms (e.g., ``bathroom'', ``kitchen''), multiple ingredients (``red apple'', ``red potato'' and ``red onion''), and multiple requirements (e.g., the red onion needs to be sliced). 
At each time step, a set of admissible actions are provided, and the player has to select an action from this set. 
After executing an action, the environment returns the observation for the next time step, the reward, and indicates whether the episode has been terminated (``Done''). 
If the agent collects an ingredient, or correctly prepares an ingredient, it will receive a positive reward from the environment, otherwise zero.
The score denotes the episodic sum of rewards. 

\subsubsection{S1 Game1}

\begin{lstlisting}
===== ----- ===== Step 0 ===== ----- =====
Obs: [['cilantro', 'cookbook', 'part_of'], ['cilantro', 'diced', 'needs'], ['cilantro', 'fridge', 'in'], ['cilantro', 'uncut', 'is'], ['cookbook', 'counter', 'on'], ['counter', 'kitchen', 'at'], ['fridge', 'closed', 'is'], ['fridge', 'kitchen', 'at'], ['knife', 'table', 'on'], ['oven', 'kitchen', 'at'], ['player', 'kitchen', 'at'], ['stove', 'kitchen', 'at'], ['table', 'kitchen', 'at']]
Admissible actions: ['examine cookbook', 'open fridge', 'take cookbook from counter', 'take knife from table']
Choose: examine cookbook
Reward: 0|Score: 0|Done: False

===== ----- ===== Step 1 ===== ----- =====
Obs: [['cilantro', 'cookbook', 'part_of'], ['cilantro', 'diced', 'needs'], ['cilantro', 'fridge', 'in'], ['cilantro', 'uncut', 'is'], ['cookbook', 'counter', 'on'], ['counter', 'kitchen', 'at'], ['fridge', 'closed', 'is'], ['fridge', 'kitchen', 'at'], ['knife', 'table', 'on'], ['oven', 'kitchen', 'at'], ['player', 'kitchen', 'at'], ['stove', 'kitchen', 'at'], ['table', 'kitchen', 'at']]
Admissible actions: ['examine cookbook', 'open fridge', 'take cookbook from counter', 'take knife from table']
Choose: open fridge
Reward: 0|Score: 0|Done: False

===== ----- ===== Step 2 ===== ----- =====
Obs: [['cilantro', 'cookbook', 'part_of'], ['cilantro', 'diced', 'needs'], ['cilantro', 'fridge', 'in'], ['cilantro', 'uncut', 'is'], ['cookbook', 'counter', 'on'], ['counter', 'kitchen', 'at'], ['fridge', 'kitchen', 'at'], ['fridge', 'open', 'is'], ['knife', 'table', 'on'], ['oven', 'kitchen', 'at'], ['player', 'kitchen', 'at'], ['stove', 'kitchen', 'at'], ['table', 'kitchen', 'at']]
Admissible actions: ['close fridge', 'examine cookbook', 'take cilantro from fridge', 'take cookbook from counter', 'take knife from table']
Choose: take cilantro from fridge
Reward: 1|Score: 1|Done: False

===== ----- ===== Step 3 ===== ----- =====
Obs: [['cilantro', 'cookbook', 'part_of'], ['cilantro', 'diced', 'needs'], ['cilantro', 'player', 'in'], ['cilantro', 'uncut', 'is'], ['cookbook', 'counter', 'on'], ['counter', 'kitchen', 'at'], ['fridge', 'kitchen', 'at'], ['fridge', 'open', 'is'], ['knife', 'table', 'on'], ['oven', 'kitchen', 'at'], ['player', 'kitchen', 'at'], ['stove', 'kitchen', 'at'], ['table', 'kitchen', 'at']]
Admissible actions: ['close fridge', 'cook cilantro with oven', 'cook cilantro with stove', 'drop cilantro', 'eat cilantro', 'examine cookbook', 'insert cilantro into fridge', 'prepare meal', 'put cilantro on counter', 'put cilantro on stove', 'put cilantro on table', 'take cookbook from counter', 'take knife from table']
Choose: take knife from table
Reward: 0|Score: 1|Done: False

===== ----- ===== Step 4 ===== ----- =====
Obs: [['cilantro', 'cookbook', 'part_of'], ['cilantro', 'diced', 'needs'], ['cilantro', 'player', 'in'], ['cilantro', 'uncut', 'is'], ['cookbook', 'counter', 'on'], ['counter', 'kitchen', 'at'], ['fridge', 'kitchen', 'at'], ['fridge', 'open', 'is'], ['knife', 'player', 'in'], ['oven', 'kitchen', 'at'], ['player', 'kitchen', 'at'], ['stove', 'kitchen', 'at'], ['table', 'kitchen', 'at']]
Admissible actions: ['chop cilantro with knife', 'close fridge', 'cook cilantro with oven', 'cook cilantro with stove', 'dice cilantro with knife', 'drop cilantro', 'drop knife', 'eat cilantro', 'examine cookbook', 'insert cilantro into fridge', 'insert knife into fridge', 'prepare meal', 'put cilantro on counter', 'put cilantro on stove', 'put cilantro on table', 'put knife on counter', 'put knife on stove', 'put knife on table', 'slice cilantro with knife', 'take cookbook from counter']
Choose: dice cilantro with knife
Reward: 1|Score: 2|Done: False

===== ----- ===== Step 5 ===== ----- =====
Obs: [['cilantro', 'cookbook', 'part_of'], ['cilantro', 'diced', 'is'], ['cilantro', 'diced', 'needs'], ['cilantro', 'player', 'in'], ['cookbook', 'counter', 'on'], ['counter', 'kitchen', 'at'], ['fridge', 'kitchen', 'at'], ['fridge', 'open', 'is'], ['knife', 'player', 'in'], ['oven', 'kitchen', 'at'], ['player', 'kitchen', 'at'], ['stove', 'kitchen', 'at'], ['table', 'kitchen', 'at']]
Admissible actions: ['close fridge', 'cook cilantro with oven', 'cook cilantro with stove', 'drop cilantro', 'drop knife', 'eat cilantro', 'examine cookbook', 'insert cilantro into fridge', 'insert knife into fridge', 'prepare meal', 'put cilantro on counter', 'put cilantro on stove', 'put cilantro on table', 'put knife on counter', 'put knife on stove', 'put knife on table', 'take cookbook from counter']
Choose: drop knife
Reward: 0|Score: 2|Done: False

===== ----- ===== Step 6 ===== ----- =====
Obs: [['cilantro', 'cookbook', 'part_of'], ['cilantro', 'diced', 'is'], ['cilantro', 'diced', 'needs'], ['cilantro', 'player', 'in'], ['cookbook', 'counter', 'on'], ['counter', 'kitchen', 'at'], ['fridge', 'kitchen', 'at'], ['fridge', 'open', 'is'], ['knife', 'kitchen', 'at'], ['oven', 'kitchen', 'at'], ['player', 'kitchen', 'at'], ['stove', 'kitchen', 'at'], ['table', 'kitchen', 'at']]
Admissible actions: ['close fridge', 'cook cilantro with oven', 'cook cilantro with stove', 'drop cilantro', 'eat cilantro', 'examine cookbook', 'insert cilantro into fridge', 'prepare meal', 'put cilantro on counter', 'put cilantro on stove', 'put cilantro on table', 'take cookbook from counter', 'take knife']
Choose: prepare meal
Reward: 1|Score: 3|Done: False

===== ----- ===== Step 7 ===== ----- =====
Obs: [['cilantro', 'cookbook', 'part_of'], ['cilantro', 'diced', 'is'], ['cilantro', 'diced', 'needs'], ['cookbook', 'counter', 'on'], ['counter', 'kitchen', 'at'], ['fridge', 'kitchen', 'at'], ['fridge', 'open', 'is'], ['knife', 'kitchen', 'at'], ['meal', 'player', 'in'], ['meal', 'raw', 'is'], ['oven', 'kitchen', 'at'], ['player', 'kitchen', 'at'], ['stove', 'kitchen', 'at'], ['table', 'kitchen', 'at']]
Admissible actions: ['close fridge', 'cook meal with oven', 'cook meal with stove', 'drop meal', 'eat meal', 'examine cookbook', 'insert meal into fridge', 'put meal on counter', 'put meal on stove', 'put meal on table', 'take cookbook from counter', 'take knife']
Choose: eat meal
Reward: 1|Score: 4|Done: True

===== ----- ===== ----- ===== ----- =====
Final obs: [['cilantro', 'cookbook', 'part_of'], ['cilantro', 'diced', 'is'], ['cilantro', 'diced', 'needs'], ['cookbook', 'counter', 'on'], ['counter', 'kitchen', 'at'], ['fridge', 'kitchen', 'at'], ['fridge', 'open', 'is'], ['knife', 'kitchen', 'at'], ['meal', 'consumed', 'is'], ['meal', 'raw', 'is'], ['oven', 'kitchen', 'at'], ['player', 'kitchen', 'at'], ['stove', 'kitchen', 'at'], ['table', 'kitchen', 'at']]

\end{lstlisting}

\subsubsection{S4 Game1}



\end{document}